\pdfoutput=1

\documentclass[11pt,dvipsnames]{article}

\usepackage[final]{emnlp2021}

\usepackage{times}
\usepackage{latexsym}

\usepackage[T1]{fontenc}

\usepackage[utf8]{inputenc}

\usepackage{microtype}
\usepackage{tabularx}
\usepackage{booktabs}
\usepackage{array}
\usepackage{amsmath}
\usepackage{amsfonts} 
\usepackage{hyperref}
\usepackage{xcolor}
\usepackage{color, colortbl}
\usepackage{adjustbox}
\usepackage{graphicx} 
\usepackage{subcaption}

\usepackage{xspace}
\newcommand{\model}{{\textsc{MirrorWiC}}\xspace}
\newcommand{\mb}{{\textsc{MirrorBert}}\xspace}

\newcommand{\hr}{{\textsc{hr}}\xspace}
\newcommand{\et}{{\textsc{et}}\xspace}
\newcommand{\ja}{{\textsc{ja}}\xspace}
\newcommand{\ko}{{\textsc{ko}}\xspace}
\newcommand{\zh}{{\textsc{zh}}\xspace}
\newcommand{\ka}{{\textsc{ka}}\xspace}
\newcommand{\ar}{{\textsc{ar}}\xspace}

\usepackage{arydshln}

\usepackage{CJKutf8}
\newcommand{\ignore}[1]{}
\newcommand\blfootnote[1]{%
  \begingroup
  \renewcommand\thefootnote{}\footnote{#1}%
  \addtocounter{footnote}{-1}%
  \endgroup
}

\usepackage{cleveref}
\crefformat{section}{\S#2#1#3}
\crefformat{subsection}{\S#2#1#3}
\crefformat{subsubsection}{\S#2#1#3}
\crefrangeformat{section}{\S\S#3#1#4 to~#5#2#6}
\crefmultiformat{section}{\S\S#2#1#3}{ and~#2#1#3}{, #2#1#3}{ and~#2#1#3}
\usepackage{refstyle}
\Crefformat{figure}{#2Fig.~#1#3}
\Crefmultiformat{figure}{Figs.~#2#1#3}{ and~#2#1#3}{, #2#1#3}{ and~#2#1#3}
\Crefformat{table}{#2Table~#1#3}
\Crefmultiformat{table}{Tabs.~#2#1#3}{ and~#2#1#3}{, #2#1#3}{ and~#2#1#3}
\Crefformat{equation}{#2Eq.~(#1#3)}

\Crefformat{appendix}{#2App.~\S#1#3}

\title{\model: On Eliciting Word-in-Context Representations \\from Pretrained Language Models}

\usepackage{tipa}

\author{Qianchu Liu$^{\ast}$, Fangyu Liu$^{\ast}$, Nigel Collier, Anna Korhonen, Ivan Vuli\'c \\
Language Technology Lab, TAL, University of Cambridge \\
\texttt{\{ql261, fl399, nhc30, alk23, iv250\}@cam.ac.uk}
}

\begin{document}
\maketitle
\begin{abstract}
Recent work indicated that pretrained language models (PLMs) such as BERT and RoBERTa can be transformed into effective sentence and word encoders even via simple self-supervised techniques. Inspired by this line of work, in this paper we propose a fully unsupervised approach to improving word-in-context (WiC) representations in PLMs, achieved via a simple and efficient WiC-targeted fine-tuning procedure: \model. The proposed method leverages only raw texts sampled from Wikipedia, assuming no sense-annotated data, and learns context-aware word representations within a standard contrastive learning setup. We experiment with a series of standard and comprehensive WiC benchmarks across multiple languages. Our proposed \textit{fully unsupervised} \model models obtain substantial gains over off-the-shelf PLMs across all monolingual, multilingual and cross-lingual setups. Moreover, on some standard WiC benchmarks, \model is even on-par with supervised models fine-tuned with in-task data and sense labels. 
\blfootnote{$^*$Equal contribution.} 
\end{abstract}

\section{Introduction}\label{sec:intro}

Pretrained Language Models (PLMs) such as BERT \cite{devlin2019bert} and RoBERTa \cite{liu2019roberta} provide dynamic contextual representations; they induce token-level lexical representations that capture the impact of the word's context on its embedding. Recent studies have assessed the PLMs by probing into their off-the-shelf representation/feature space \cite{gari-soler-etal-2019-comparison,wiedemann2019does,Reif:2019neurips,soler2021let}. While off-the-shelf PLMs already offer a useful contextualised lexical semantic space, their contextualised representation spaces suffer from instability and anisotropy \cite{mickus-etal-2020-mean,pedinotti-lenci-2020-dont}. As a consequence, they usually fall far behind the performance of the same PLM fine-tuned with (i) sense annotations \cite{hadiwinoto-etal-2019-improved,blevins-zettlemoyer-2020-moving} or (ii) external (e.g., WordNet) knowledge \cite{levine-etal-2020-sensebert}.

\begin{figure}[t]
    \centering
    \includegraphics[width=0.999\linewidth]{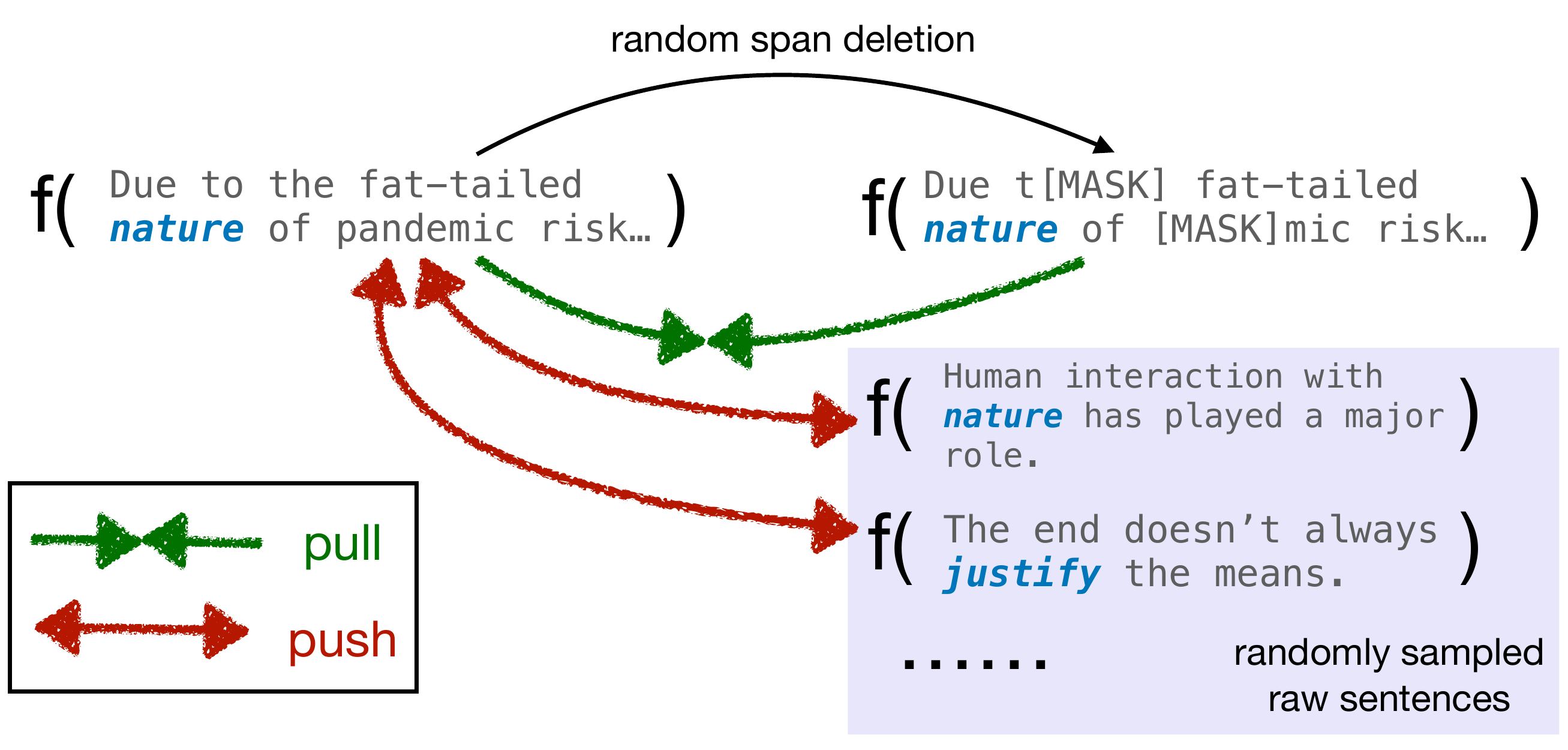}
    \caption{An illustrative overview of the \model method, based on contrastive learning, for eliciting better word-in-context (WiC) representations from pretrained language models. We augment a randomly selected WiC instance with random span masking and apply dropout to the hidden states to create two slightly different representations of the base instance. These two representations form a positive pair for contrastive fine-tuning. During fine-tuning, we pull the representations of each positive pair closer together, while at the same time pushing away representations of other WiC instances, serving as negative examples.}
    \label{fig:overview}
\end{figure}

However, PLMs have been shown to actually store more lexical and sentence-level information than what can be directly extracted from their off-the-shelf variants. In simple words, this knowledge must be `unlocked' or exposed via additional adaptive fine-tuning \cite{Ruder:2021blog}. For instance, while off-the-shelf PLMs are not directly effective as universal sentence encoders, it is possible to convert them into such encoders through supervised \cite{reimers2019sentence,Feng:2020labse,liu2020self} or self-supervised fine-tuning \cite{carlsson2021semantic,liu2021fast,gao2021simcse} based on the \textit{contrastive learning} paradigm.  

The fundamental limitation of extracting contextual features/representations directly from the layers of the off-the-shelf PLMs is the mismatch between their (pre)training objectives and the feature extraction method. In other words, the contextual representations, typically extracted as the averages over the top four layers of a base PLM \cite{liu2020towards,soler2021let}, can be seen as a by-product of training a language model, and are not directly optimised for contextual sensitivity. Inspired by the previous work on adaptive fine-tuning for word and sentence representations \cite{liu2021fast}, we propose a simple self-supervised technique termed \model: it  \textit{rewires} input PLMs to provide improved word-in-context (WiC) representations. 

Unlike prior work on fine-tuning towards improving WiC representations, our \model procedure disposes of any sense labels, annotated task data, and any external knowledge, and elicits improved WiC representations from PLMs in a \textit{fully unsupervised} way. We design a contrastive learning framework that directly optimises the contextual representations (i.e., the top four hidden layers of the input PLM) that are also the feature space at inference time; see Figure~\ref{fig:overview} and Table~\ref{tab:train_inference_consistency}. \model relies on the sets of positive and negative pairs, where the positive pairs are created by pairing an input sequence (which contains a target word) with its slightly altered variant. This altered sequence is obtained via random span masking and the resulting representations for this pair are further altered by dropout. The negative pairs are then simply the same or different word's contextual representations in a different context; Figure~\ref{fig:overview}. These pairs for fine-tuning are mined from raw Wikipedia sentences. To understand why \model works so well, we provide ablation studies on the design choices (including dropout rate, random span masking, etc.) and layer-wise analyses and visualisation on \model's effects on embedding properties such as isotropy. 






\vspace{2.0mm}
\noindent \textbf{Contributions.} \textbf{1)} We present a simple yet extremely effective unsupervised \model technique for eliciting contextual lexical knowledge. \textbf{2)} Our experiments on a range of English, multilingual, and cross-lingual context-sensitive lexical benchmarks demonstrate that \model achieves consistent and substantial improvements over different baseline PLMs, indicating its robustness and wide applicability. \textbf{3)} We offer extensive analyses and additional insights into the inner workings of \model, and its impact on the contextual representation space. We release our code at {\small \url{github.com/cambridgeltl/MirrorWiC}}. 


\section{Related Work and Background}\label{sec:rw}

\noindent \textbf{Word-in-Context Representations.} Modelling context influence on lexical meaning and creating context-aware word representations is a long-standing research goal in lexical semantics. One direction is to create discrete sense embeddings according to a fixed sense inventory such as WordNet. These embeddings can be created from the attributes in the sense inventory such as glosses \cite{chen-etal-2014-unified} or from the knowledge structure \cite{camacho2016nasari}. We point to \citet{camacho2018word} for a thorough survey on sense embeddings. Such sense representations require a fixed and discrete sense inventory and might not be sensitive enough to the the dynamic and fluid nature of contextual changes. 

More recently, PLMs provide dynamic and continuous contextual representations, not tied to predefined sense inventories, computed as a function of both the target word and its context. The use of PLMs has resulted in further progress on a range of context-aware evaluation benchmarks \cite{pilehvar2019wic,Wang:2019superglue,raganato-etal-2020-xl}. A body of work has aimed to enrich context-aware and sense information in the PLMs by injecting such knowledge (e.g., sense annotations from predefined sense inventories) at pretraining stage \cite{levine-etal-2020-sensebert} or during inference \cite{loureiro-jorge-2019-language}. Other work has attempted at combining/ensembling multiple contextualised and static type-level embeddings to refine the contextualised representation space \citep{liu2020towards,xu2020improving}.

\vspace{2.0mm}
\noindent \textbf{Inducing Text Representations from PLMs via Self-Supervision.} Recently, there has been growing interest in learning completely unsupervised sentence representations from PLMs using contrastive learning techniques \citep{carlsson2021semantic,liu2021fast,gao2021simcse,yan2021consert,kim-etal-2021-self,zhang-etal-2021-bootstrapped}. Similar to the supervised approaches such as Sentence-BERT \cite{reimers-gurevych-2019-sentence} or SapBERT \cite{liu2020self}, the idea is to transform an input PLM into an effective sentence encoder via additional fine-tuning. During self-supervised contrastive fine-tuning, the model learns from identical or automatically modified text sequences (treated as positive examples), and regards different sentences as negative pairs. \mb \citep{liu2021fast} is a general self-supervised contrastive fine-tuning framework that transforms off-the-shelf PLMs into effective word and sentence encoders. Our proposed \model method can be seen as an extension of \mb, now focused on eliciting improved word-in-context representations and context-sensitive lexical tasks.

\begin{table*}[!t]
    \centering
    \small
    \begin{tabular}{lll}
    \toprule
model & representations fine-tuned & representations extracted \\
\midrule
off-the-shelf PLMs & (\texttt{[CLS]} +) language modelling head & word token average (top four layers) \\
\mb & \texttt{[CLS]}/mean-pooling & \texttt{[CLS]}/mean-pooling\\
\model &  word token average (top four layers) & word token average (top four layers)\\
    \bottomrule
    \end{tabular}
    \caption{\model benefits from the consistency of representations at (i) fine-tuning and (ii) feature extraction and inference: both are focused on word-in-context (WiC) representations.}
    \label{tab:train_inference_consistency}
\end{table*}
\begin{table}[!t]
    \centering
    \def\arraystretch{0.97}
    \setlength{\tabcolsep}{1.2pt}
    \resizebox{1.0\columnwidth}{!}{
    \begin{tabular}{cl}
    \toprule
    \multicolumn{2}{l}{Step 1: Automatic dataset creation for WiC fine-tuning} \\
    \midrule
        $({\color{Blue}x_1},{\color{magenta}y_1})$ &  ({\color{Blue}\textit{Due to the fat-tailed \textbf{nature} of pandemic risk, \ldots}}, {\color{magenta}$1$}) \\
        $({\color{Blue}\overline{x}_1},{\color{magenta}\overline{y}_1)}$ &   ({\color{Blue}\textit{Due to the fat-tailed \textbf{nature} of pandemic risk, \dots},} {\color{magenta}$1$})\\
        \ldots & \ldots \\
        $({\color{Blue}x_i},{\color{magenta}y_i})$ &  ({\color{Blue}\textit{Human interaction with \textbf{nature} has played a major role.}}, {\color{magenta}$i$}) \\
        $({\color{Blue}\overline{x}_i},{\color{magenta}\overline{y}_i})$ &  ({\color{Blue}\textit{Human interaction with \textbf{nature} has played a major role.}}, {\color{magenta}$i$}) \\
        \ldots & \ldots \\
        $({\color{Blue}x_N}, {\color{magenta}y_N})$ &  ({\color{Blue}\textit{The end doesn't always \textbf{justify} the means.}}, {\color{magenta}$N$}) \\
        $({\color{Blue}\overline{x}_N}, {\color{magenta}\overline{y}_N})$ &  ({\color{Blue}\textit{The end doesn't always \textbf{justify} the means.}}, {\color{magenta}$N$}) \\
        \midrule
    \multicolumn{2}{l}{Step 2: Random span masking} \\
        \midrule
                $({\color{Blue}x_1}, {\color{magenta}y_1})$ &  ({\color{Blue}\textit{Due to the fat-tailed \textbf{nature} of pandemic risk, \dots}}, {\color{magenta}$1$}) \\
        $({\color{Blue}\overline{x}_1}, {\color{magenta}\overline{y}_1})$ &   ({\color{Blue}\textit{Due \_e fat-tailed \textbf{nature} of pandemic \_ \ldots}}, {\color{magenta}$1$})\\
        \ldots & \ldots \\
        $({\color{Blue}x_i}, {\color{magenta}y_i})$ &  ({\color{Blue}\textit{Human interaction with \textbf{nature} has played a major role.}}, {\color{magenta}$i$}) \\
        $({\color{Blue}\overline{x}_i},{\color{magenta}\overline{y}_i)}$ & ({\color{Blue}\textit{Human intera\_ with \textbf{nature} has play\_major role.}}, {\color{magenta}$i$}) \\
        \ldots & \ldots \\
        $({\color{Blue}x_N},{\color{magenta}y_N})$ &  ({\color{Blue}\textit{the end does not always \textbf{justify} The means.}},{\color{magenta}$N$}) \\
        $({\color{Blue}\overline{x}_N},{\color{magenta}\overline{y}_N)}$ &  ({\color{Blue}\textit{The end does\_always \textbf{justify} \_eans.}}, {\color{magenta}$N$}) \\
    \bottomrule
    \end{tabular}}
    \caption{Upper: the automatically generated labelled dataset for fine-tuning PLMs towards learning better word-in-context representations. \textbf{Bold} denotes the target word. Lower: data augmentation via random span masking. `\_' denotes the `[MASK]' token.}
    \label{tab:data}
    
\end{table}

\section{\model: Methodology}
\label{sec:method}

\noindent \textbf{Baseline WiC Representations.} 
Prior work directly extracts word-in-context representations from the parameters of the off-the-shelf PLMs. The most effective (empirically validated) strategy is 1) averaging the representations from the top four PLM's layers, and 2) then taking either the first constituent subword from the PLM's vocabulary to represent the target word, or further averaging the representations of the word's constituent subwords \citep{liu2020towards,soler2021let}. 


\subsection{Self-Supervised WiC Fine-Tuning} 
We hypothesise that it is possible to convert the input PLM into an improved WiC encoder through adaptive (self-supervised) fine-tuning. Given a set of raw sentences without labels, how do we tune the PLMs to further expose their word-in-context knowledge?   Inspired by \mb \citep{liu2021fast}, we apply a self-supervised contrastive learning scheme to elicit better word-in-context representations. We fine-tune the input PLM by contrasting the representations of different word-in-context pairs while pulling representations of a self-duplicated word-in-context pair closer in the representation space (see Figure~\ref{fig:overview}).

\vspace{2.0mm}
\noindent \textbf{Data Creation.} 
Given a set of $N$ non-duplicated sentences, we randomly select a word in each sentence as the target word: i.e., the sentences become a set of `word-in-context instances'. We then follow \mb and generate a labelled dataset by duplicating each instance in the set and assigning identical labels to identical instances and different labels to different word-in-contexts (\Cref{tab:data}, upper half): $\mathcal{D} = \{(x_1, y_1), (\overline{x}_1, \overline{y}_1), \ldots, (x_N, y_N), (\overline{x}_N, \overline{y}_N)\}$, where $\forall i=1,\ldots N$, it holds $x_i = \overline{x}_i, y_i = \overline{y}_i$.

\vspace{2.0mm}
\noindent \textbf{Data Augmentation.} We further follow \mb to create a slightly altered (or augmented) `view' of the same text sequence: we randomly replace a span of text with `[MASK]'\footnote{Or `<MASK>' for input to RoBERTa.} in all duplicated examples. There is a fundamental difference to \mb where such `random span masking' technique is applied on sentences; for word-in-context, we keep the target word intact (otherwise the semantics changes drastically) and randomly replace a span of length $K$ on \textit{both sides} of the target word; see \Cref{tab:data} (lower half).
Besides random span masking, the dropout modules in the Transformer layers also slightly and randomly alter the representations of each word-in-context instance. They serve as another source of data augmentation to further perturb the word-in-context representations. After both input space augmentation (random span masking) and feature space augmentation (dropout layers embedded in the Transformer layers), the resulting embeddings of even a positive pair will be slightly different.\footnote{Note that random span masking is applied on only one instance of each duplicated pair, while the dropouts are applied to all instances.}


\vspace{2.0mm}
\noindent \textbf{Contrastive Fine-Tuning.} Following the feature extraction procedure from off-the-shelf PLMs, we compute the average of hidden states from the PLM's top four layers, and then take the average of all token(s) that correspond to the target word, as the word-in-context representation. Let $f(\cdot)$ denote the encoder which outputs such WiC representation. We leverage InfoNCE \citep{oord2018representation} to cluster/attract the positive pairs together and push away the negative pairs in the embedding space:

\vspace{-2mm}
{\footnotesize
\begin{equation}
    \mathcal{L} = -\sum_{i=1}^{N}\log\frac{\exp(\cos(f(x_i), f(\overline{x}_i))/\tau)}{\displaystyle \sum_{x_j\in \mathcal{N}_i}\exp(\cos(f(x_i), f(x_j))/\tau)}.
    \label{eq:infonce}
\end{equation}
}%
\noindent where $\tau$ is a tunable temperature; $\mathcal{N}_i$ denotes all negatives of $x_i$, which includes all $x_j,\overline{x}_j$ where $i\neq j$ in the current data batch (i.e., $|\mathcal{N}_i| = N - 2$). Intuitively, the numerator is the similarity of the self-duplicated pair (a positive pair) and the denominator is the sum of the similarities between $x_i$ and all other strings besides $\overline{x}_i$ (negative pairs). 

For positive pairs, though one sequence in the pair is slightly altered via random span masking and the representations go through dropout, the encoding function $f(\cdot)$ should learn an invariant mapping and reconstruct the correct semantics from the noise \cite{liu2021fast}. Most negative examples contain different target words and different contexts (e.g., $x_1$ and $x_N$ in \Cref{tab:data}). Naturally, such pairs are of different meanings and the model should produce different representations.Note that it is possible to also have the same target word appearing in different contexts as a negative pair (e.g., $x_1$ and $x_i$ in \Cref{tab:data}). If the pair indeed has very different semantics (of a different sense), then pushing them apart is actually desirable. However, even if the items in the pair happen to have similar meanings, our learning objective still instructs the model to push them away from each other. Our rationale and decision here are based on the following: 
(1) Such \textit{false} negative pairs can act as a regularisation; and (2) in essence, one could argue that all distinct word-in-context instances have slightly different meanings since sense is a continuous function of word and context.


\section{Experimental Setup}
\label{sec:exp}

\noindent \textbf{WiC Evaluation.} 
We evaluate \model on a range of context-sensitive lexical semantic tasks in monolingual English settings, as well as in multilingual and cross-lingual settings.

For English, we evaluate on two similarity-based tasks: \textit{Usim} and \textit{CoSimLex}; two word-in-context classification tasks: \textit{WiC} and \textit{WiC-TSV}; and one-shot Word Sense Disambiguation (WSD). {Usim} \citep{erk2013measuring} measures the similarity between two instances of the same word occurring in two different sentential contexts. {CoSimLex} \cite{armendariz2020cosimlex} measures the change in similarity between two different words appearing in two different contexts: paragraphs. We follow the standard evaluation protocol, computing the cosine similarity of the contextual word representations and comparing them against human-elicited scores via Spearman's rank correlation ($\rho$). 

The WiC classification task \citep{pilehvar2019wic} challenges a model to make a binary decision on whether or not the same target word has the same meaning in two different contexts. The {WiC-TSV} (TSV) task \citep{breit-etal-2021-wic} extends the original \textit{WiC} to multiple domains with three different subtasks. In TSV-1, the task is to decide if the intended sense of the target word in the context matches the target sense described by the definition. In TSV-2, the model must identify if the intended sense (in the context) is the hyponym of the provided hypernyms. TSV-3 combines the previous two subtasks (see \citet{breit-etal-2021-wic} for further details). 

The WSD task \cite{navigli2009word,Raganato:2017eacl} requires a system to select the correct label for a given target word in context from a candidate set of all possible meanings for this target word. To evaluate the feature space of the models in WSD, we create a one-shot setting where we provide one context example\footnote{The context examples are taken from WordNet entries. If a sense does not contain context, we reformat the definition as '<target word> means ...' as the target word's context.} per label and perform nearest neighbour search over contextual word representations from the candidate labels. We directly test the models on the concatenated \texttt{ALL} test set from \citet{Raganato:2017eacl} without access to training and development data. 

 We also perform multilingual and cross-lingual evaluation on \textit{XL-WiC} \citep{raganato-etal-2020-xl} and \textit{AM2iCo} \citep{liu2021am2ico}. XL-WiC provides WiC-style evaluations in multiple languages. AM2iCo extends XL-WiC to lower-resource languages, adds more difficult adversarial examples, and enables cross-lingual evaluations. For brevity, we show results for four typologically diverse languages both from XL-WiC (\zh, \ko, \hr, \et); and four languages in AM2iCo (\zh, \ka, \ja, \ar).\footnote{\zh: Mandarin Chinese, \ko: Korean, \hr: Croatian, \et: Estonian, \ka: Georgian, \ja: Japanese, \ar: Arabic.}

For WiC, TSV, XL-WiC and AM2iCo, our main experiments follow the unsupervised method from \citet{pilehvar2019wic}: we compute cosine similarity between the contextual word representations in each pair, and search for a threshold to divide true (i.e., same meaning) and false instances on the development set in each task.\footnote{We add templates in each TSV subtask: `[target word] means <definition>' (TSV-1); `[target word] is a kind of <hypernym>' (TSV-2) and `[target word] is a kind of <hypernym> and means <definition>' (TSV-3). We then compute similarity based on the contextual representations of the target words in these templates. This results in an unsupervised approach which is more effective than the approach from prior work \citep{breit-etal-2021-wic}, where cosine similarity is computed on definition/hypernym embeddings.} We report accuracy scores in the main paper, while additional area-under-curve (AUC) scores are available in \Cref{sec:auc}.

\vspace{2.0mm}
\noindent \textbf{Underlying PLMs.}
We experiment with several standard input PLMs for English, but we remind the reader that the \model framework is applicable with a wide range of PLMs: \textbf{1)} BERT \cite{devlin2019bert} as a standard choice for WiC representation learning and evaluation \cite{raganato-etal-2020-xl}; \textbf{2)} RoBERTa \cite{liu2019roberta} as an optimised and improved PLM; and \textbf{3)} DeBERTa \cite{he2020deberta} as a more recent PLM that achieves state-of-the-art results in a range of natural language understanding tasks \cite{Wang:2019superglue}.\footnote{DeBERTa extends the standard BERT architecture by incorporating two novel techniques: disentangled attention that encodes a word's content and position separately, and an enhanced masked decoder that incorporates absolute position for predicting masked tokens during masked language modelling.} For all non-English experiments, unless noted otherwise, we rely on multilingual BERT (mBERT) as the underlying PLM (see \Cref{sec:encoder}).

\vspace{2.0mm}
\noindent \textbf{Fine-Tuning Details.} We largely follow the \mb fine-tuning setup \cite{liu2021fast}, using 10k sentences randomly drawn from Wikipedia as the \model fine-tuning corpus. For monolingual models, we sample 10k sentences from the corresponding Wikipedia of that language. For cross-lingual models, we sample 5k sentences from English Wikipedia and 5k from Wikipedia of each target language. We train all models with AdamW \cite{loshchilov2018decoupled} with a learning rate of \texttt{2e-5} for 1 epoch. The $\tau$ in \Cref{eq:infonce} is set to $0.04$.  We set $K$ (random span masking rate) to $10$, $0$ and $1$ for BERT, RoBERTa and DeBERTa respectively. The respective dropout rates are $0.4$, $0.3$ and $0.3$ for BERT, RoBERTa and DeBERTa. All hyper-parameters are tuned on the development set of WiC and kept unchanged for all other experiments. We refer the reader to the Appendix (\Cref{Table:search_space}) for a full listing of hyperparameters along with their search space.





\section{Results and Discussion}
\label{sec:results}
\begin{table*}[t!]
\footnotesize
\setlength{\tabcolsep}{3.0pt}
\centering
\begin{tabular}{lcccccccccccccccccccccccccccccc}
\toprule
  model$\downarrow$, dataset$\rightarrow$ & Usim ($\rho$) & WiC (acc) &	TSV-1 (acc) & TSV-2 (acc) &  TSV-3 (acc) & CoSimLex ($\rho$) & One-shot WSD (acc)\\ 
  \midrule  

Sentence-BERT &23.57& 61.91 & 62.46 &59.64 &62.72&-& 42.63\\
  
  \mb &23.21 & 64.10 & 66.32 & 64.78 & 66.32 & - & 44.93\\
 \midrule 
    
BERT & 54.52 &	68.49 & 61.69 & 60.66 & 61.95	& 76.2 & 52.90 \\

 \rowcolor{cyan!10}
  + \model & 61.82 & \textbf{71.94} & 69.15 & 66.06 & 68.38  & 77.41 & 57.10\\
  \cmidrule(l){2-9}
RoBERTa& 50.25&	66.77&	55.52&	56.55&	57.58& 75.64 &51.38\\
\rowcolor{cyan!10}
  + \model & 57.95& 71.15	&69.92&	67.60&	71.70& 77.27 & 56.51\\
 \cmidrule(l){2-9}
 DeBERTa& 54.77& 66.14 & 59.38 &59.89  & 60.41 &72.06& 53.99 \\
  \rowcolor{cyan!10}
 + \model & \textbf{62.79} & 71.78 & \textbf{70.95} & \textbf{67.86} & \textbf{71.20} & \textbf{77.70}& \textbf{59.02}\\
 
\bottomrule
\end{tabular}
\caption{Results across a collection of context-aware lexical semantic tasks in English. \label{table:main}}
\end{table*}

\subsection{Main Results: Evaluation on English}
\label{ss:resen}

The main results are provided in \Cref{table:main} and \Cref{table:sup}. Most notably, we observe consistent and substantial gains over all unsupervised baselines, including the off-the-shelf PLMs without \model fine-tuning. While the underlying PLMs, as suggested by prior work \cite{soler2021let}, do encode a wealth of sense-related knowledge, that knowledge can be further exposed via the proposed context-aware \model fine-tuning procedure.

\begin{figure*}[!t] 
\begin{subfigure}{0.4\textwidth}
\includegraphics[width=\linewidth]{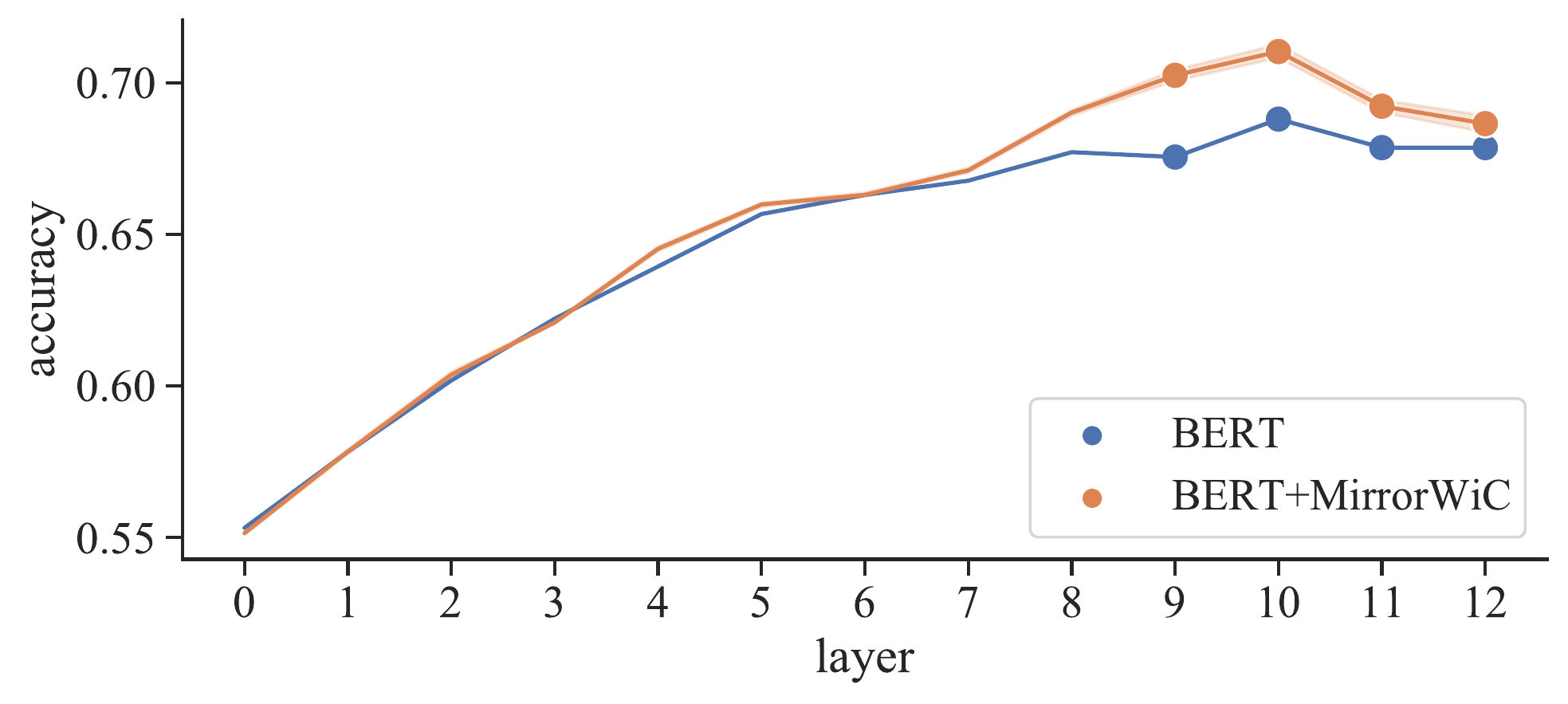}
\caption{{BERT layer-wise accuracy; WiC dev }} \label{fig:wic_bert}
\end{subfigure}\hspace*{\fill}
\begin{subfigure}{0.4\textwidth}
\includegraphics[width=\linewidth]{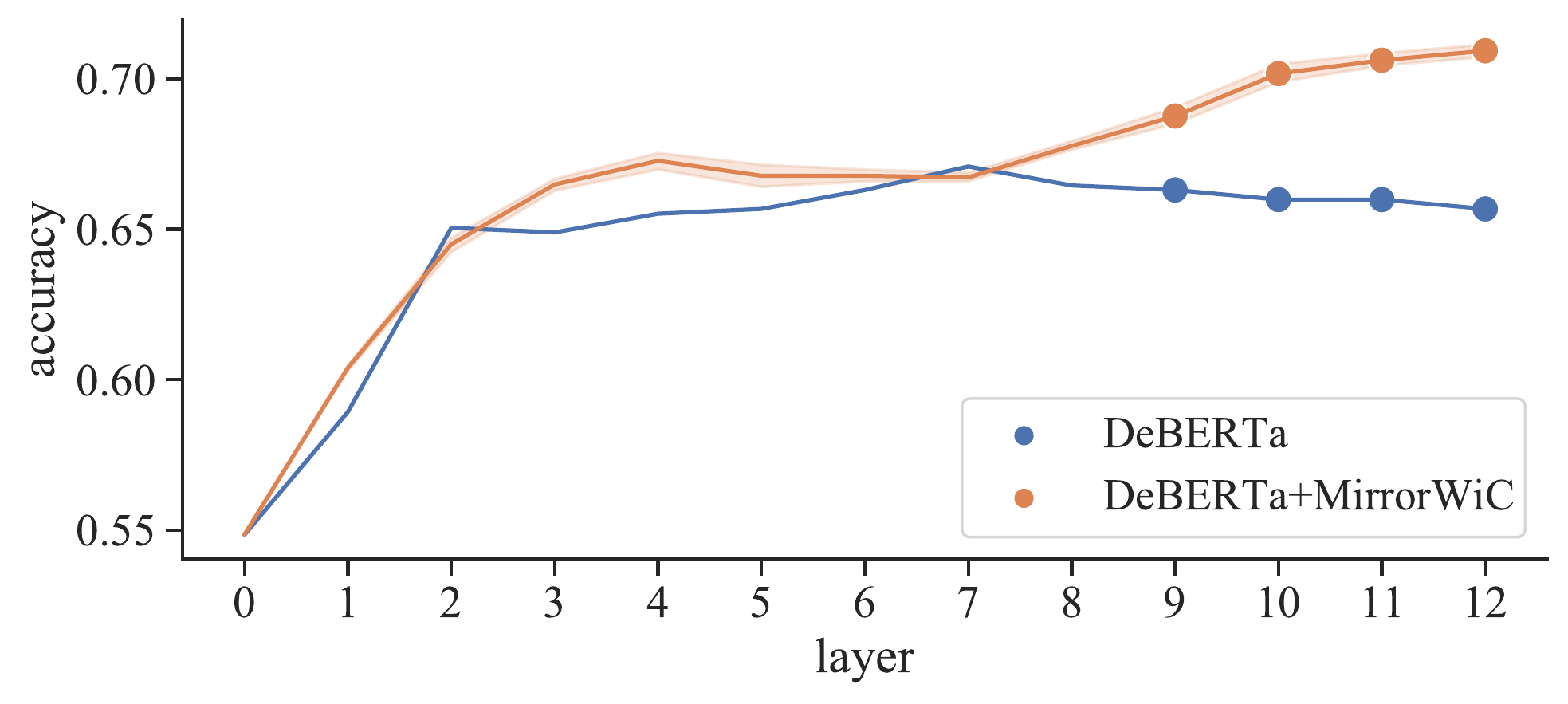}
\caption{{DeBERTa layer-wise accuracy; WiC dev }} \label{fig:wic_deberta}
\end{subfigure}
\medskip

\begin{subfigure}{0.4\textwidth}
\includegraphics[width=\linewidth]{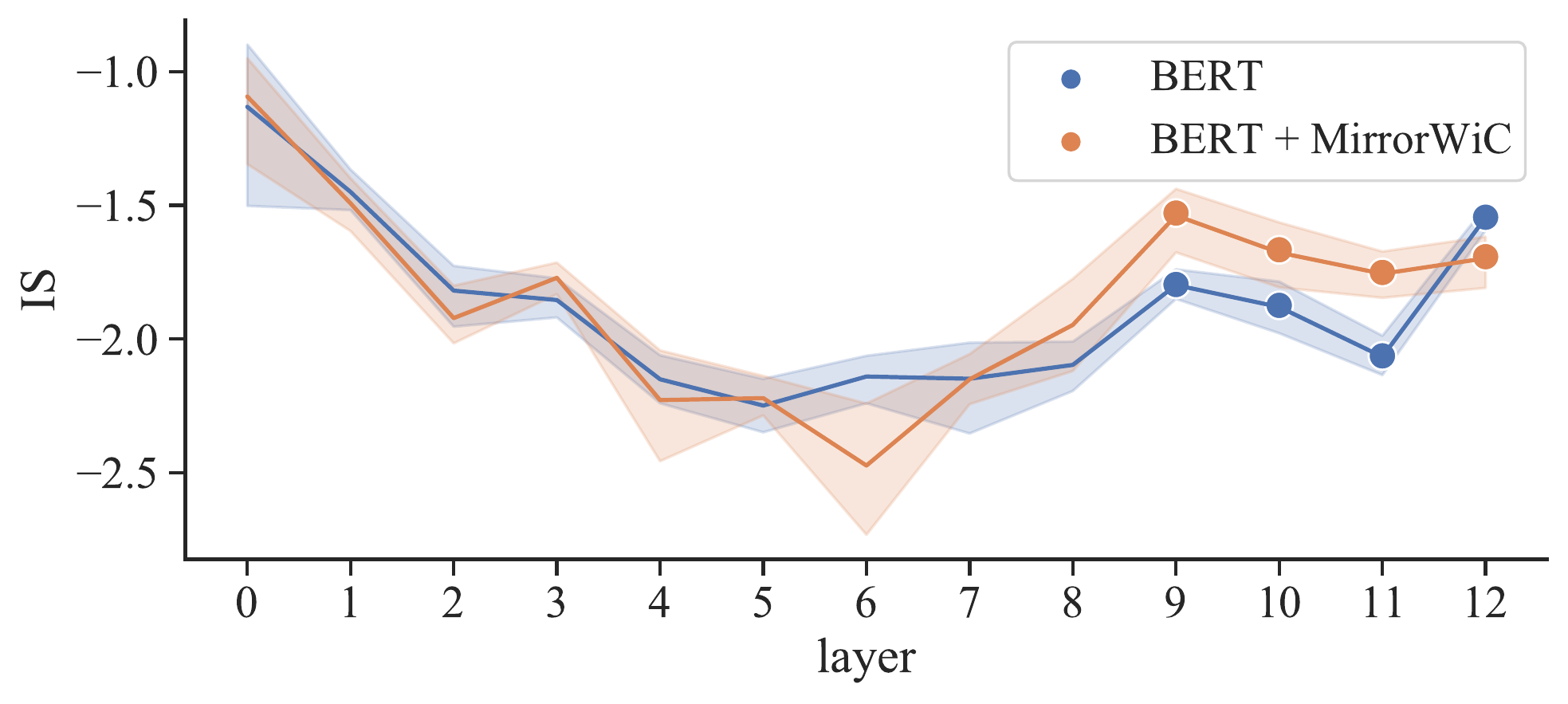}
\caption{BERT isotropy (higher is better)} \label{fig:iso_bert}
\end{subfigure}\hspace*{\fill}
\begin{subfigure}{0.4\textwidth}
\includegraphics[width=\linewidth]{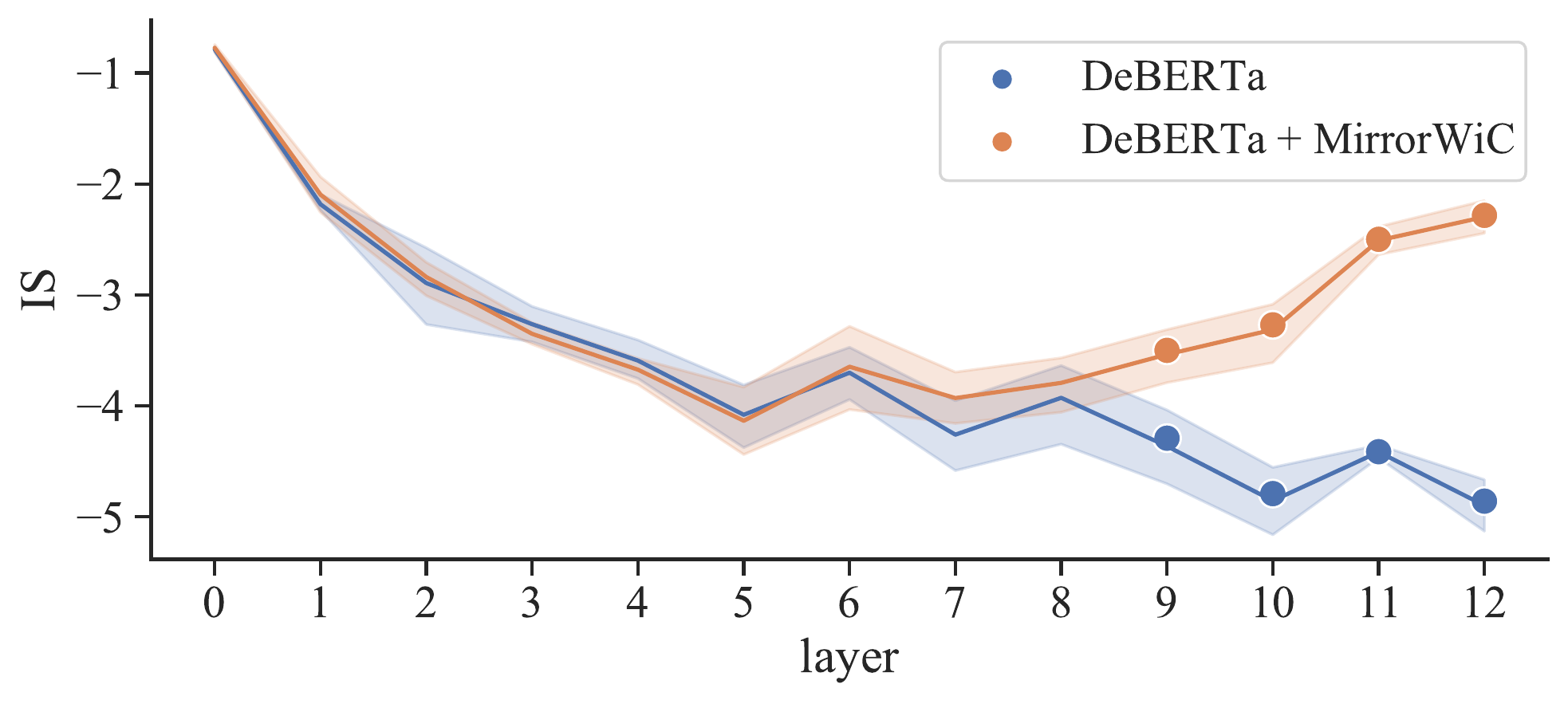}
\caption{DeBERTa isotropy (higher is better)} \label{fig:iso_deberta}
\end{subfigure}

\medskip
\begin{subfigure}{0.4\textwidth}
\includegraphics[width=\linewidth]{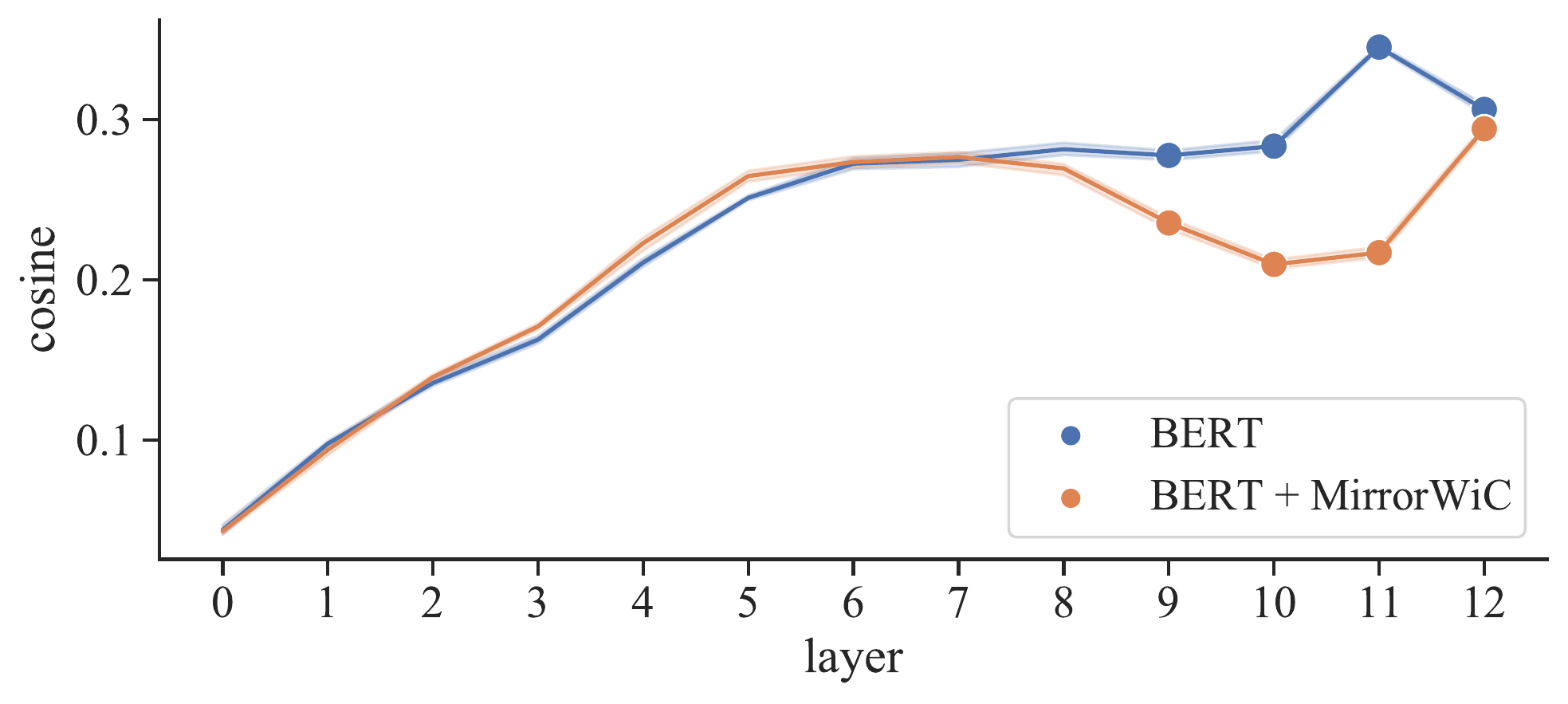}
\caption{BERT random-word cosine similarity} \label{fig:ran_bert}
\end{subfigure}\hspace*{\fill}
\begin{subfigure}{0.4\textwidth}
\includegraphics[width=\linewidth]{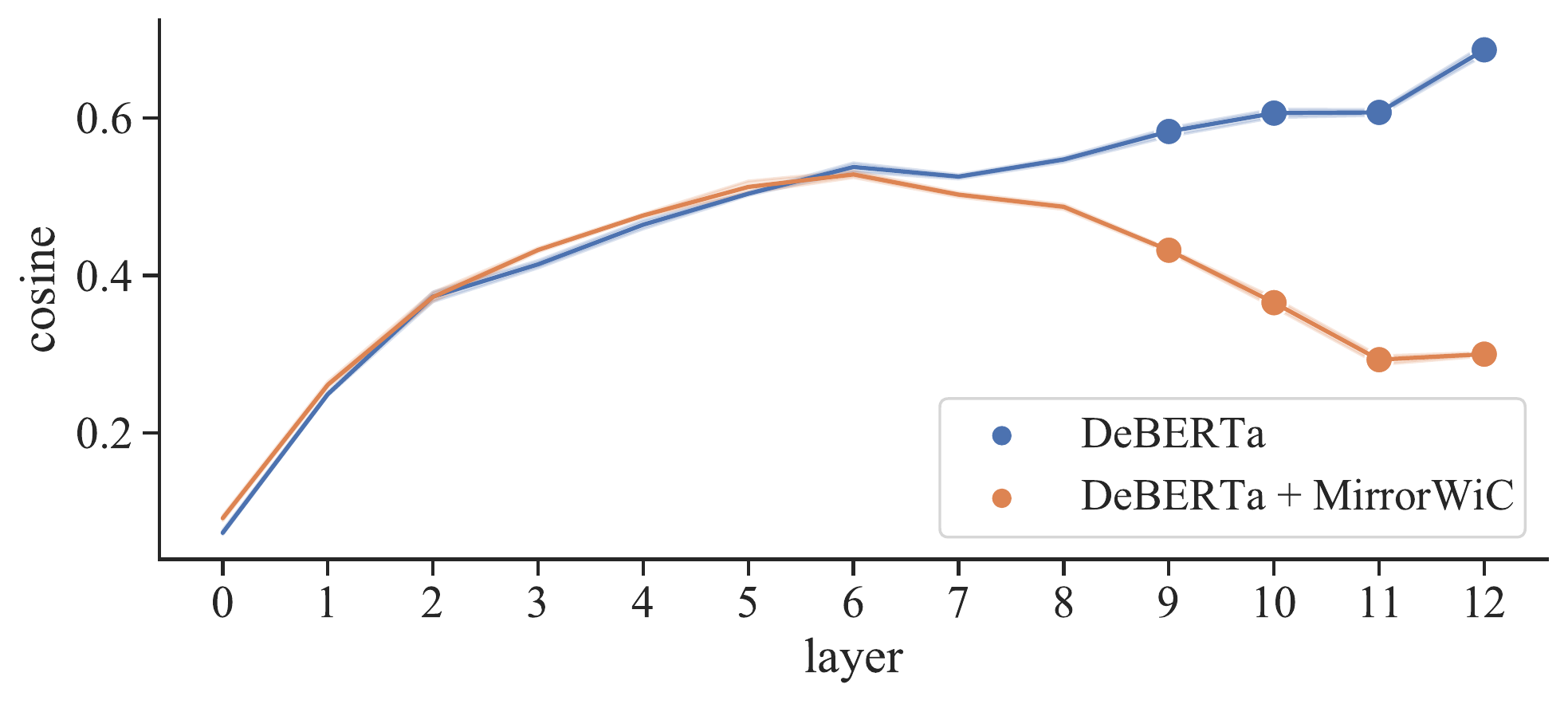}
\caption{DeBERTa random-word cosine similarity} \label{fig:ran_deberta}
\end{subfigure}

\medskip
\begin{subfigure}{0.4\textwidth}
\includegraphics[width=\linewidth]{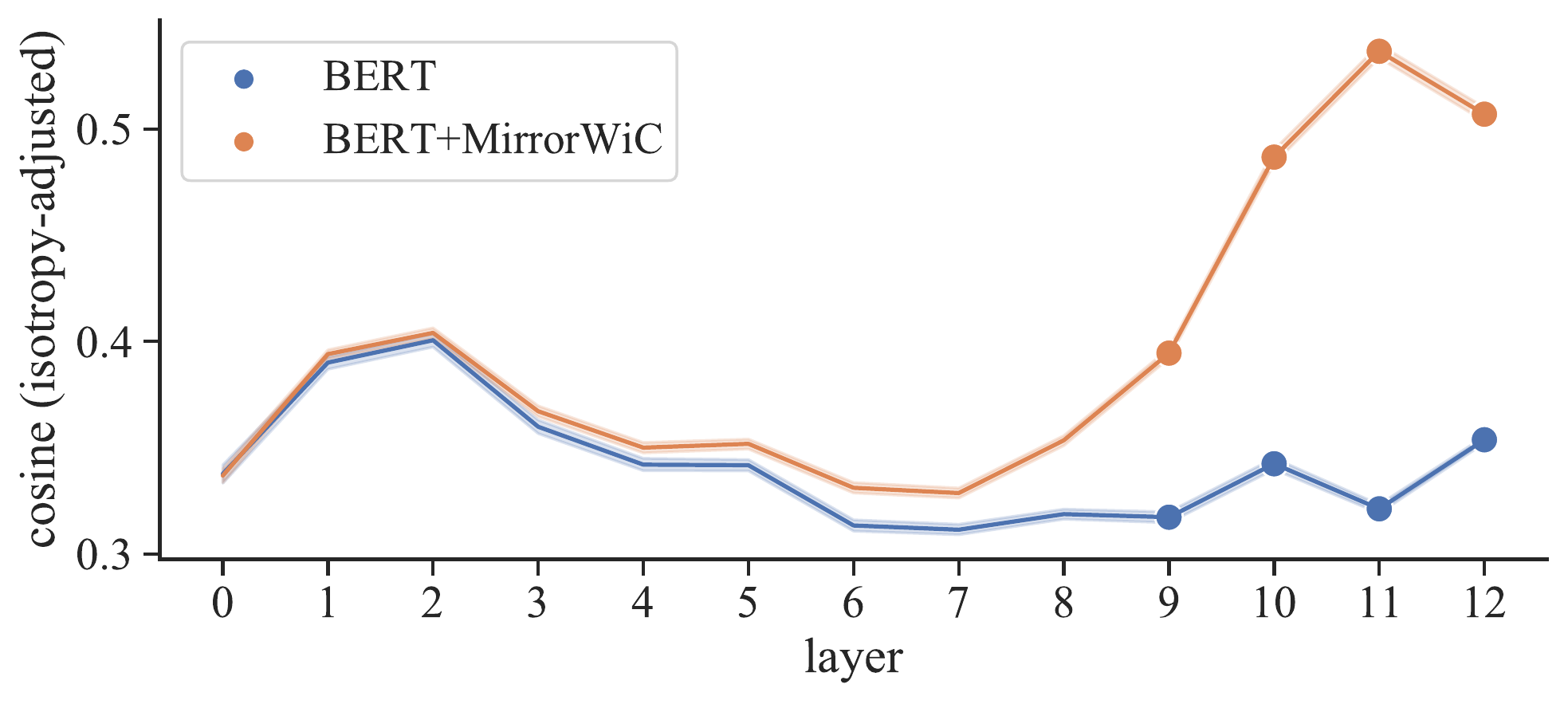}
\caption{BERT intra-sentence cosine similarity} \label{fig:intra_bert}
\end{subfigure}\hspace*{\fill}
\begin{subfigure}{0.4\textwidth}
\includegraphics[width=\linewidth]{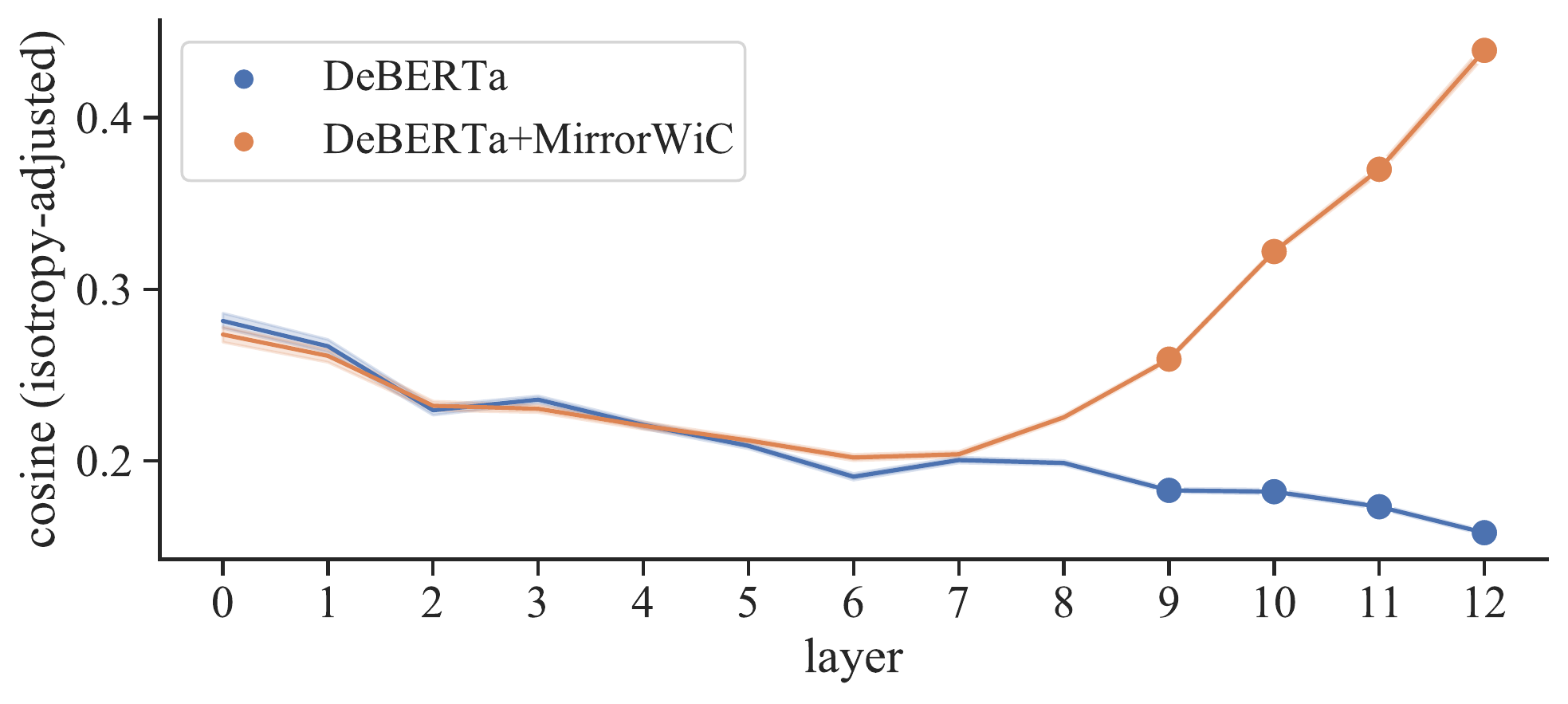}
\caption{DeBERTa intra-sentence cosine similarity} \label{fig:intra_deberta}
\end{subfigure}

\caption{Layer-wise analyses of BERT (left column) and DeBERTA (right column) before and after applying \model.  The first row (a,b) shows the model performance and can be linked to the isotropy analysis (the middle two rows: c,d,e,f) and contextualisation analysis (the last row: g,b). Task performance correlates strongly with isotropy and contextualisation changes especially in the last four layers (highlighted with dots); shade=variance.} \label{fig:analysis}

\end{figure*}


\vspace{2.0mm}
\noindent \textbf{Impact of the Underlying PLM (\Cref{table:main}).} \model is effective with BERT, RoBERTa and DeBERTa.
DeBERTa+\model yields larger gains, and even results in the highest absolute scores on average. In other words, a seemingly `weaker' off-the-shelf PLM under the naive feature extraction baseline (DeBERTa) is transformed into the best-performing WiC encoder after the \model procedure. This hints at the necessity to unlock the input PLM's 'task solving potential' through adaptive fine-tuning. 


\vspace{2.0mm}
\noindent \textbf{Comparison with Sentence Encoders (\Cref{table:main}).}  We also probe how modelling the sentences (without knowing which target word the context is describing) performs on the evaluation tasks. In particular, we evaluate the standard 'go-to' sentence encoder Sentence-BERT \citep{reimers-gurevych-2019-sentence}, and the original \mb \citep{liu2021fast}. We find that \model, with its direct focus on word-in-context representations and WiC-oriented fine-tuning, substantially outperforms the two sentence encoders. The finding validates our hypothesis that naively applying sentence encoders is not sufficient for context-aware lexical semantic tasks. While the two sentence encoders do provide competitive performance in WiC-style tasks, their performance decreases drastically on Usim. This further indicates that the fine-grained similarity-based Usim evaluation requires a more accurate and subtler contextual lexical semantic ability than the binary classification in WiC. 

\vspace{2.0mm}
\noindent \textbf{Comparison with Supervised WiC Methods (\Cref{table:sup}).} The scores reveal that the unsupervised BERT + \model variant can even outperform the supervised model (fine-tuned with labelled in-task data) in the WiC task. The results on TSV indicate that the gap between the unsupervised BERT-based approach to the supervised performance is much reduced: from the $\sim$10\% gap to only $\sim$2\% in all three TSV tasks when \model is applied. 



\subsection{Multilingual and Cross-Lingual Results}

The results are summarised in \Cref{table:xl}. Notably, we observe that the effectiveness of \model is not tied to English, and extends to other languages. We observe consistent improvements with the underlying PLMs monolingually pretrained in other languages, as well as with the multilingually pretrained mBERT. The gains on XL-WiC are more pronounced than on the more difficult AM2iCo benchmark. By design AM2iCo is a more challenging benchmark, and additional external knowledge injection might be necessary to improve the results further; unlike XL-WiC, AM2ico requires the models to understand the cross-lingual correspondence of mostly entity names that occur less frequently. 

\begin{table}[t!]
\small
\setlength{\tabcolsep}{4.2pt}
\centering
\begin{tabular}{llccccccccccccccccccccccccccccc}
\toprule
  model$\downarrow$, dataset$\rightarrow$ &  WiC   &	TSV-1 &	TSV-2 &  TSV-3  \\ 
  \midrule  
    
BERT &	65.85 & 65.08 & 62.09& 63.16\\
 \rowcolor{cyan!10}
  + \model & \textbf{69.64} & 73.66	&  69.83 & 73.73 \\
\midrule
 task-supervised BERT& 69.00 & \textbf{75.30} & \textbf{71.40} & \textbf{76.60}  \\
\bottomrule
\end{tabular}
\caption{BERT+\model versus supervised BERT-based methods on the test sets of English WiC-style tasks. The supervised variant on WiC is replicated from \citet{Wang:2019superglue}. The supervised results on TSV are taken from \citet{breit-etal-2021-wic}.}
\label{table:sup}
\end{table}
\begin{table}[!t]
\centering
{\small
\begin{tabular}{lccccccccccccccccccccccccccccccccc}
\toprule
 XL-WiC & \zh* & \ko* & \hr & \et\\
\midrule
BERT & 73.74 &	68.41 &	61.10 & 57.06 \\
 \rowcolor{cyan!10}
  + \model & \textbf{75.70} & \textbf{72.26} & \textbf{67.32} &	\textbf{61.43} \\
  \midrule
  \midrule
  
   AM2iCo & \zh &	\ka & \ja &	\ar \\ 
  \midrule
 BERT &  63.80 &	59.90 &	64.10 &	60.60  \\
 \rowcolor{cyan!10}
+ \model &   \textbf{64.60} & \textbf{61.00} & \textbf{64.70} &	\textbf{63.90} \\
\bottomrule
\end{tabular}
}%
\caption{Results (test set accuracy) on multilingual and cross-lingual word-in-context tasks. We use mBERT as the underlying PLM for all the languages except for \zh* and \ko* (in XL-WiC) where their monolingual BERT models were used. \label{table:xl}}
\end{table}

\subsection{Further Discussion and Analyses}
\label{ss:further}

\noindent \textbf{Layer-wise Performance (\Cref{fig:wic_bert,fig:wic_deberta}).} The figures reveal that the success of \model is attributed to the performance gains achieved in the last four layers of the fine-tuned PLMs. This is expected as these four layers are exactly what we optimise in the \model procedure. This also confirms our hypothesis that matching training and inference representations helps adapt and elicit word-in-context knowledge from the PLMs. 

\vspace{2.0mm}
\noindent \textbf{Isotropy (\Cref{fig:iso_bert,fig:iso_deberta}).} 
As empirically validated in prior work on sentence representations \cite{gao2021simcse,liu2021fast}, contrastive fine-tuning reshapes the embedding space geometry towards more isotropic representations, which in turn has a positive impact on semantic similarity tasks. We now examine whether the same `isotropy-increasing' effect is achieved with \model. To this end, we leverage a quantitative isotropy score (IS), proposed in prior work \citep{arora2016latent,mu2017all},\footnote{The same metric is used for measuring isotropy of contextual word representations by \citet{rajaee-pilehvar-2021-cluster}.} and defined as:

\vspace{-1mm}
{\footnotesize
\begin{equation}
  \text{IS} (\mathcal{V}) =  \log \left(\frac{\min_{\mathbf{c}\in \mathcal{C}} \sum_{\mathbf{v}\in \mathcal{V}}\exp(\mathbf{c}^\top\mathbf{v})}{\max_{\mathbf{c}\in \mathcal{C}} \sum_{\mathbf{v}\in \mathcal{V}}\exp(\mathbf{c}^\top\mathbf{v})} \right)
\end{equation}
}%
where $\mathcal{V}$ is the set of vectors, $\mathcal{C}$ is the set of all possible unit vectors in the embedding space (i.e., $\{\mathbf{c} : |\mathbf{c}|=1\}$). Practically, $\mathcal{C}$ is approximated by the eigenvector set of $\mathbf{V}^\top\mathbf{V}$ ($\mathbf{V}$ is the stacked embeddings of $\mathcal{V}$). The larger the IS value, the more isotropic an embedding space is.\footnote{We randomly sample 10k sentences from English Wikipedia as $\mathcal{V}$. We compute the average word-in-context embeddings for all words in each sentence and then compute the IS value. We repeat the process for five times to reduce the randomness introduced in sampling.} 

As seen in \Cref{fig:iso_bert} and \Cref{fig:iso_deberta}, both BERT and DeBERTa create more isotropic embedding spaces in general in the last four layers after \model training. Note that DeBERTa's space isotropy is able to benefit more from \model, which also explains its large gains in the end tasks.

It is also possible to assess isotropy by simply looking at the cosine similarity of random words \cite{ethayarajh-2019-contextual}. We calculate word representations in each layer as the average of the word's contextual representations from Wikipedia. We then take five random samples of 200 random words and compute pair-wise similarity. We take the average of the similarity scores in each sample with variance reported in \Cref{fig:ran_bert} and \Cref{fig:ran_deberta}. The results confirm the trend: the last four layers with \model exhibit much lower random word cosine similarities than the off-the-shelf PLM.

\begin{table*}[th]
{\footnotesize
\begin{tabularx}{\textwidth}{p{0.33\textwidth}p{0.27\textwidth}p{0.1\textwidth}p{0.12\textwidth}p{0.05\textwidth}}
\toprule
 {\bf Word-in-context 1} & {\bf Word-in-context 2}&{\bf \textsc{BERT}} & {\bf \textsc{+\model}} &Gold\\ 
\midrule

\textit{\textbf{Spend} money.}	& \textit{He \textbf{spends} far more on gambling than he does on living proper.} &	-0.0850~(F) &0.2327~(T)  &T\\
\midrule[0.05pt]
\textit{That toaster can make wonderful \textbf{toasts}.} &	\textit{I ate a piece of \textbf{toast} for breakfast.} &	0.0160~(F) &0.3234~(T)& T \\
\midrule
\textit{War is \textbf{hell}.} & \textit{The \textbf{hell} of battle.} &-0.0403~(F)&0.2378~(T)	&F\\
\midrule[0.05pt]
\textit{\textbf{Ease} the pain in your legs.} &	\textit{The pain \textbf{eased} overnight.} &	0.0157~(F)&0.2873~(T)&F\\
\bottomrule

\end{tabularx}
}
\caption{Examples of changed cosine similarity scores (isotropy-adjusted) after \model; English WiC (dev). \label{table:wic_error}}
\end{table*}

\begin{figure*}[h]
    \centering
    \includegraphics[width=0.93\linewidth]{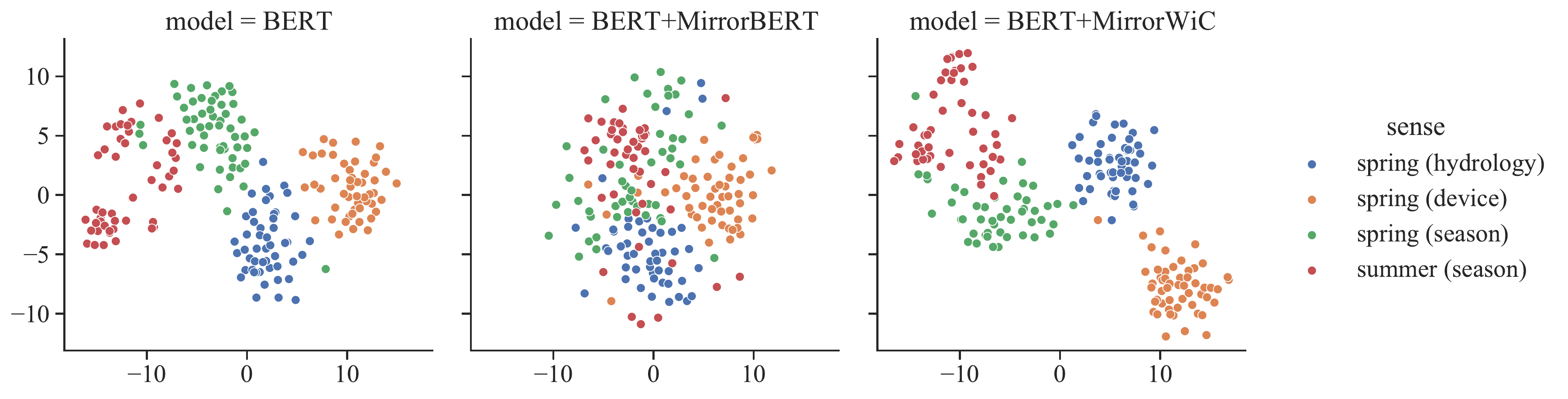}
    \caption{t-SNE embedding visualisation of different senses of \textit{spring} and \textit{summer} under different models.}
    \label{fig:tsne_spring}
\end{figure*}

\vspace{2.0mm}
\noindent \textbf{Intra-Sentence Similarity (\Cref{fig:intra_bert,fig:intra_deberta}).} 
As a measure of \textit{contextualisation}, we follow \citet{ethayarajh-2019-contextual}, and define intra-sentence similarity as each word's similarity to its context. The context is computed as the mean vector of all the word representations in the sentence. The scores are isotropy-adjusted \ignore{Flora: consistent with the figure} by substracting the intra-sentence similarity scores by the random word similarity in each layer, see \cite{ethayarajh-2019-contextual}. For both BERT and DeBERTa, we can see that the last four layers become more contextualised after applying \model: they encode more information about the context as the contextual word representations become much more similar to its context in the top Transformer layers than in the base PLM. This increased contextualisation could explain why \model gives better performance in the context-sensitive lexical semantic tasks.

\vspace{2.0mm}
\noindent \textbf{Error Analysis (\Cref{table:wic_error}).}
Conducting an error analysis of BERT before and after \model on the WiC dev set, we observe that 94 instances changed their labels, among which 58 are \model correcting the original predictions. In 43 out of 58 cases, \model is producing more TRUE positives. The examples with the largest similarity changes are provided in the upper half of \Cref{table:wic_error}. 
For the 36 cases where \model changes the originally correct predictions to the wrong prediction, 29 are false positives; see the lower half of \Cref{table:wic_error}. We manually inspect these cases and find that the distinctions between the two contexts are usually too fine-grained to tell even for humans. For instance, it seems acceptable to align with the \model's (incorrect) predictions for \textbf{\textit{hell}} and \textbf{\textit{ease}} in the two examples in \Cref{table:wic_error}. 



\vspace{2.0mm}
\noindent \textbf{Visualising the Embedding Space (\Cref{fig:tsne_spring}).}
Contextualised embeddings for an ambiguous word (\textit{spring}) with off-the-shelf BERT, \mb and \model are visualised in \Cref{fig:tsne_spring} (sense labels from Wikipedia). While \model maintains the sense clusters from BERT and teases apart the different senses even more, \mb exhibits no clear sense distinctions. This shows a fundamental difference between \model and \mb: \mb is insensitive to the target word, and directly applying it to context-sensitive lexical tasks yields subpar performance.  

\subsection{Ablation Study}
An ablation study is conducted on English WiC (dev). Foreshadowing, the dropout rate and the layer averaging strategy are the two most important factors for \model to be effective. 

\vspace{2.0mm}
\noindent \textbf{Dropout and Random Span Masking (\Cref{tab:dropout,tab:random_erasing}).}
The \model performance is most sensitive to the dropout rate; it requires larger dropout rates (0.3 for DeBERTa and 0.4 for BERT) than \mb (0.1 dropout). This may be related to the different levels of granularity. Sentence meanings can largely change with even slight differences in context: therefore, positive sentence pairs for \mb are required to be very similar. Word-in-context meaning can tolerate larger contextual differences: larger dropout rates are thus preferable with \model to create positive pairs with more distinct representations. Random span masking is less crucial than the dropout rate, and gives only slight gains (\Cref{tab:random_erasing}). 

\vspace{2.0mm}
\noindent \textbf{Layer Averaging Strategy (\Cref{tab:layer}).}
Averaging across all layers of the PLM is suboptimal for WiC representations, and the strategy of averaging only over the last four layers is indeed the optimal one for BERT. However, DeBERTa reaches its peak when averaging over the last 2 layers. Our findings corroborate those from previous studies which report that contextualised information is usually stored in higher layers \cite{ethayarajh-2019-contextual,soler2021let}, and the bulk of decontextualised information is stored in lower layers \cite{vulic-etal-2020-probing}.

\begin{figure}[!t]
    \centering
    \begin{subfigure}{0.24\textwidth}
    \includegraphics[width=\linewidth]{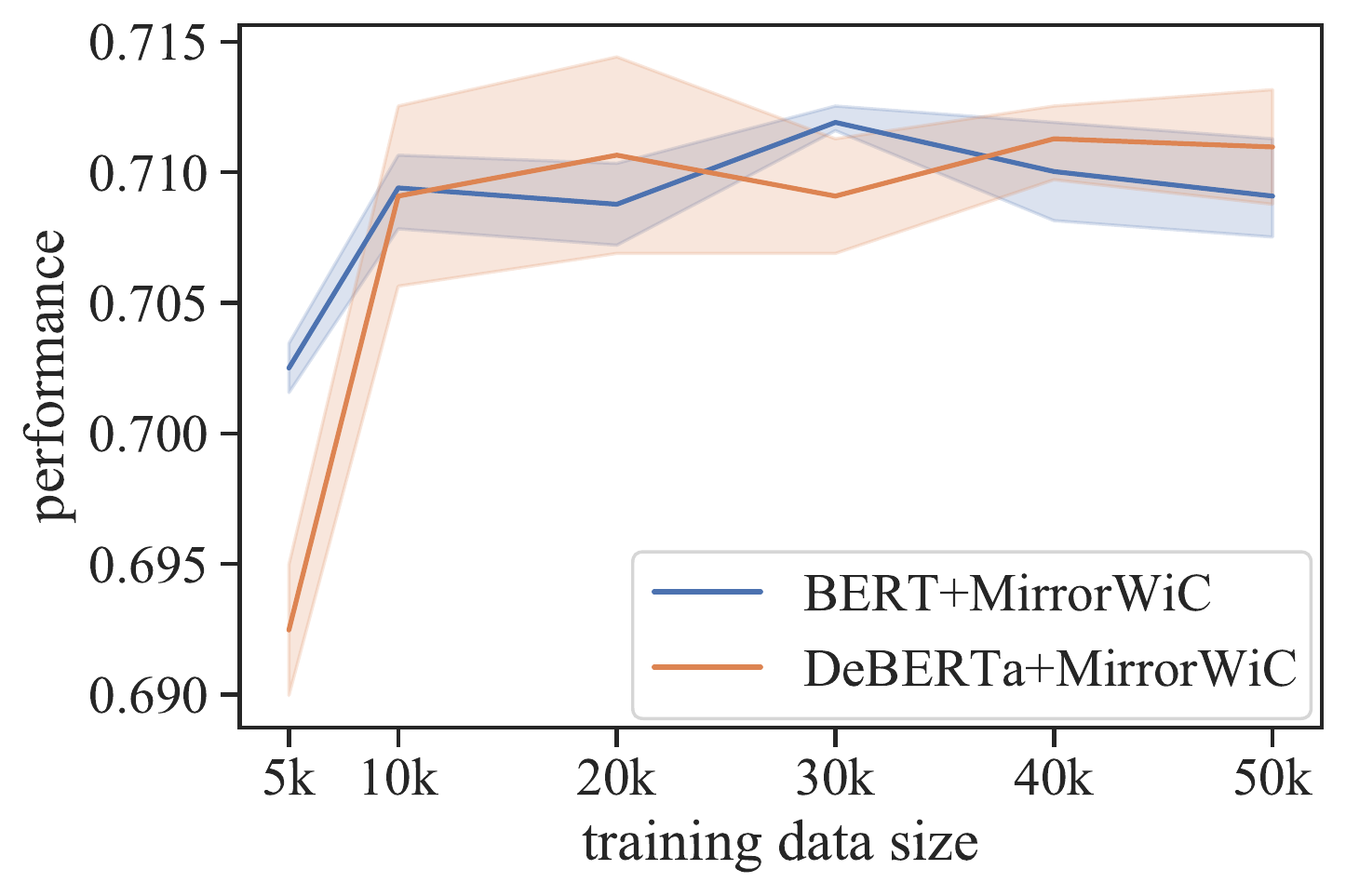}
    \caption{WiC (dev)}
    \end{subfigure}\hspace*{\fill}
    \medskip
    \begin{subfigure}{0.24\textwidth}
    \includegraphics[width=\linewidth]{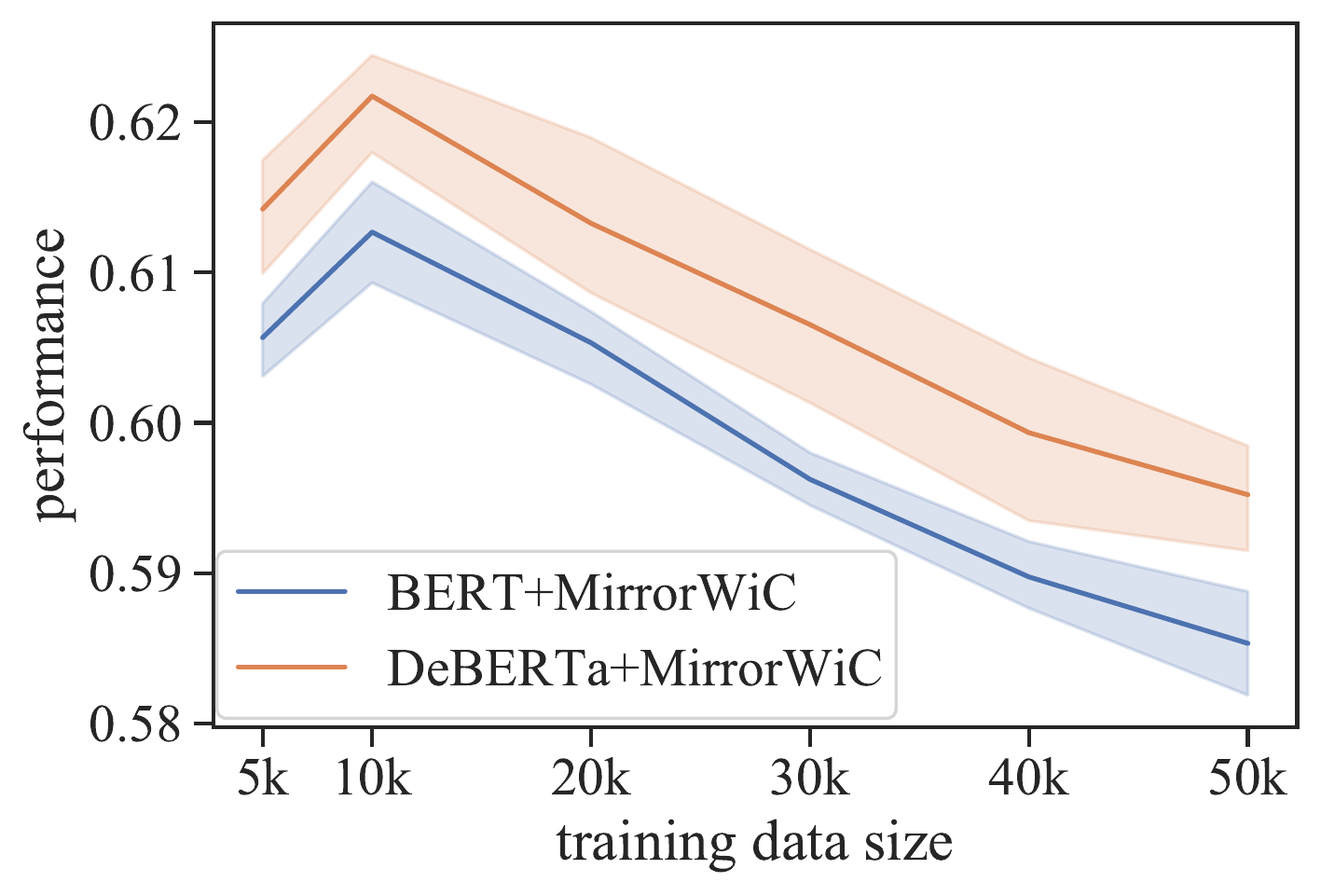}
    \caption{Usim}
    \end{subfigure}
    \caption{Impact of input size of the training data for \model. Evaluation on WiC (dev).}
    \label{fig:input_size}
\end{figure}

\vspace{2.0mm}
\noindent \textbf{Input Size (\Cref{fig:input_size}).}
As in \Cref{fig:input_size}, we show a sharp increase of performance from 5k to 10k on both Usim and WiC. While WiC maintains its performance with small fluctuation from 10k throughout to 50k, there is a clear downward slope for Usim from 10k onward. This is in line with findings in \mb, and also shows that the model does not require plenty of fine-tuning data to transform into a WiC encoder. This further confirms that the model is not so much learning new knowledge as rewiring knowledge to the surface.  


\begin{table}[!t]
\scriptsize
\setlength{\tabcolsep}{2.2pt}
\centering
\begin{tabular}{lcccccccccccccccccccccccccccccc}
\toprule
dropout rate$\rightarrow$ &  0   &	0.1 &	0.2 &  0.3 & 0.4 & 0.5 & 0.6  \\ 
  \midrule  
  BERT + \model & 68.02&	68.65&	70.21&	71.31&	{\bf 71.94}&	68.80&	68.49 \\
 DeBERTa + \model & 65.67	&69.12	 & 70.53	&{\bf 71.78}	& 67.08	&65.98	&66.30 \\
\bottomrule
\end{tabular}
\caption{Impact of dropout rate in \model.}
\label{tab:dropout}
\end{table}
\begin{table}[!t]
\scriptsize
\setlength{\tabcolsep}{4.2pt}
\centering
\begin{tabular}{llccccccccccccccccccccccccccccc}
\toprule
  model$\downarrow$, random span masking$\rightarrow$ & off & on \\ 
  \midrule  
 BERT + \model &  71.31 & 71.47$_{\uparrow0.16}$\\
 DeBERTa + \model & 71.78 & 71.94$_{\uparrow0.16}$ \\
\bottomrule
\end{tabular}
\caption{Impact of random span masking.}
\label{tab:random_erasing}
\end{table}

\begin{table}[!t]
\scriptsize
\setlength{\tabcolsep}{2.2pt}
\centering
\begin{tabular}{lcccccccccccccccccccccccccccccc}
\toprule
average last $n$ layers$\rightarrow$ &  1 & 2 &	3 &  4 & 5& 6 & 12  \\ 
  \midrule  
  BERT + \model & 68.96 & 68.80 & 70.06&	\textbf{71.94}&	70.68&	70.84&	67.71 \\
 DeBERTa + \model & 71.47&	\textbf{73.04}&	72.41	&71.78	&71.15 &70.53& 69.74\\
\bottomrule
\end{tabular}
\caption{Impact of layer averaging strategies.}
\label{tab:layer}
\end{table}





\section{Conclusion}
We proposed  \model, a fully unsupervised approach for eliciting word-in-context representations from pretrained language models (PLMs), requiring only raw sentences as input, and disposing of labelled data and sense inventories. We showed that \model is PLM-agnostic and language-agnostic, yielding substantial performance boosts in context-aware lexical semantic tasks in English, multilingual and cross-lingual setups and demonstrating that additional WiC knowledge can be exposed from the PLMs. We then delved into the inner-working of \model, demonstrating that the performance improvement strongly correlates with metrics such as isotropy score and intra-sentence word similarity. In future work, we will also look into weakly supervised approaches that combine self-supervision with external sense-related knowledge.

\section*{Acknowledgements}
We thank the three reviewers and the ACs
for their helpful feedback. We acknowledge Peterhouse College at University of Cambridge for funding Qianchu Liu's PhD, and Grace \& Thomas C.H. Chan Cambridge Scholarship for funding Fangyu Liu's PhD. The work has also been funded by the ERC Grant LEXICAL (no. 648909) and the ERC PoC Grant MultiConvAI (no. 957356) awarded to Anna Korhonen.  


\bibliography{anthology,custom}

\begin{thebibliography}{45}
\expandafter\ifx\csname natexlab\endcsname\relax\def\natexlab#1{#1}\fi

\bibitem[{Armendariz et~al.(2020)Armendariz, Purver, Ul{\v{c}}ar, Pollak,
  Ljube{\v{s}}i{\'c}, and Granroth-Wilding}]{armendariz2020cosimlex}
Carlos~Santos Armendariz, Matthew Purver, Matej Ul{\v{c}}ar, Senja Pollak,
  Nikola Ljube{\v{s}}i{\'c}, and Mark Granroth-Wilding. 2020.
\newblock \href {https://www.aclweb.org/anthology/2020.lrec-1.720}
  {{C}o{S}im{L}ex: A resource for evaluating graded word similarity in
  context}.
\newblock In \emph{Proceedings of the 12th Language Resources and Evaluation
  Conference}, pages 5878--5886, Marseille, France. European Language Resources
  Association.

\bibitem[{Arora et~al.(2016)Arora, Li, Liang, Ma, and
  Risteski}]{arora2016latent}
Sanjeev Arora, Yuanzhi Li, Yingyu Liang, Tengyu Ma, and Andrej Risteski. 2016.
\newblock \href {https://doi.org/10.1162/tacl_a_00106} {A latent variable model
  approach to {PMI}-based word embeddings}.
\newblock \emph{Transactions of the Association for Computational Linguistics},
  4:385--399.

\bibitem[{Blevins and Zettlemoyer(2020)}]{blevins-zettlemoyer-2020-moving}
Terra Blevins and Luke Zettlemoyer. 2020.
\newblock \href {https://doi.org/10.18653/v1/2020.acl-main.95} {Moving down the
  long tail of word sense disambiguation with gloss informed bi-encoders}.
\newblock In \emph{Proceedings of the 58th Annual Meeting of the Association
  for Computational Linguistics}, pages 1006--1017, Online. Association for
  Computational Linguistics.

\bibitem[{Breit et~al.(2021)Breit, Revenko, Rezaee, Pilehvar, and
  Camacho-Collados}]{breit-etal-2021-wic}
Anna Breit, Artem Revenko, Kiamehr Rezaee, Mohammad~Taher Pilehvar, and Jose
  Camacho-Collados. 2021.
\newblock \href {https://www.aclweb.org/anthology/2021.eacl-main.140}
  {{WiC-TSV}: {A}n evaluation benchmark for target sense verification of words
  in context}.
\newblock In \emph{Proceedings of the 16th Conference of the European Chapter
  of the Association for Computational Linguistics: Main Volume}, pages
  1635--1645, Online. Association for Computational Linguistics.

\bibitem[{Camacho-Collados and Pilehvar(2018)}]{camacho2018word}
Jose Camacho-Collados and Mohammad~Taher Pilehvar. 2018.
\newblock \href {https://arxiv.org/abs/1805.04032} {From word to sense
  embeddings: A survey on vector representations of meaning}.
\newblock \emph{Journal of Artificial Intelligence Research}, 63:743--788.

\bibitem[{Camacho-Collados et~al.(2016)Camacho-Collados, Pilehvar, and
  Navigli}]{camacho2016nasari}
Jos{\'e} Camacho-Collados, Mohammad~Taher Pilehvar, and Roberto Navigli. 2016.
\newblock \href
  {https://www.sciencedirect.com/science/article/pii/S0004370216300820}
  {Nasari: Integrating explicit knowledge and corpus statistics for a
  multilingual representation of concepts and entities}.
\newblock \emph{Artificial Intelligence}, 240:36--64.

\bibitem[{Carlsson et~al.(2021)Carlsson, Gyllensten, Gogoulou, Hellqvist, and
  Sahlgren}]{carlsson2021semantic}
Fredrik Carlsson, Amaru~Cuba Gyllensten, Evangelia Gogoulou,
  Erik~Ylip{\"a}{\"a} Hellqvist, and Magnus Sahlgren. 2021.
\newblock \href {https://openreview.net/forum?id=Ov_sMNau-PF} {Semantic
  re-tuning with contrastive tension}.
\newblock In \emph{International Conference on Learning Representations}.

\bibitem[{Chen et~al.(2014)Chen, Liu, and Sun}]{chen-etal-2014-unified}
Xinxiong Chen, Zhiyuan Liu, and Maosong Sun. 2014.
\newblock \href {https://doi.org/10.3115/v1/D14-1110} {A unified model for word
  sense representation and disambiguation}.
\newblock In \emph{Proceedings of the 2014 Conference on Empirical Methods in
  Natural Language Processing ({EMNLP})}, pages 1025--1035, Doha, Qatar.
  Association for Computational Linguistics.

\bibitem[{Devlin et~al.(2019)Devlin, Chang, Lee, and
  Toutanova}]{devlin2019bert}
Jacob Devlin, Ming-Wei Chang, Kenton Lee, and Kristina Toutanova. 2019.
\newblock \href {https://doi.org/10.18653/v1/N19-1423} {{BERT}: Pre-training of
  deep bidirectional transformers for language understanding}.
\newblock In \emph{Proceedings of the 2019 Conference of the North {A}merican
  Chapter of the Association for Computational Linguistics: Human Language
  Technologies, Volume 1 (Long and Short Papers)}, pages 4171--4186,
  Minneapolis, Minnesota. Association for Computational Linguistics.

\bibitem[{Erk et~al.(2013)Erk, McCarthy, and Gaylord}]{erk2013measuring}
Katrin Erk, Diana McCarthy, and Nicholas Gaylord. 2013.
\newblock \href {https://doi.org/10.1162/COLI_a_00142} {Measuring word meaning
  in context}.
\newblock \emph{Computational Linguistics}, 39(3):511--554.

\bibitem[{Ethayarajh(2019)}]{ethayarajh-2019-contextual}
Kawin Ethayarajh. 2019.
\newblock \href {https://doi.org/10.18653/v1/D19-1006} {How contextual are
  contextualized word representations? comparing the geometry of {BERT},
  {ELM}o, and {GPT}-2 embeddings}.
\newblock In \emph{Proceedings of the 2019 Conference on Empirical Methods in
  Natural Language Processing and the 9th International Joint Conference on
  Natural Language Processing (EMNLP-IJCNLP)}, pages 55--65, Hong Kong, China.
  Association for Computational Linguistics.

\bibitem[{Feng et~al.(2020)Feng, Yang, Cer, Arivazhagan, and
  Wang}]{Feng:2020labse}
Fangxiaoyu Feng, Yinfei Yang, Daniel Cer, Naveen Arivazhagan, and Wei Wang.
  2020.
\newblock \href {https://arxiv.org/abs/2007.01852} {Language-agnostic {BERT}
  sentence embedding}.
\newblock \emph{CoRR}, abs/2007.01852.

\bibitem[{Gao et~al.(2021)Gao, Yao, and Chen}]{gao2021simcse}
Tianyu Gao, Xingcheng Yao, and Danqi Chen. 2021.
\newblock \href {https://arxiv.org/pdf/2104.08821.pdf} {Simcse: Simple
  contrastive learning of sentence embeddings}.
\newblock \emph{arXiv preprint arXiv:2104.08821}.

\bibitem[{Gar{\'\i}~Soler and Apidianaki(2021)}]{soler2021let}
Aina Gar{\'\i}~Soler and Marianna Apidianaki. 2021.
\newblock \href {https://arxiv.org/abs/2104.14694} {Let's play mono-poly: Bert
  can reveal words' polysemy level and partitionability into senses}.
\newblock \emph{Transactions of the Association for Computational Linguistics
  (TACL)}.

\bibitem[{Gar{\'\i}~Soler et~al.(2019)Gar{\'\i}~Soler, Cocos, Apidianaki, and
  Callison-Burch}]{gari-soler-etal-2019-comparison}
Aina Gar{\'\i}~Soler, Anne Cocos, Marianna Apidianaki, and Chris
  Callison-Burch. 2019.
\newblock \href {https://doi.org/10.18653/v1/W19-0423} {A comparison of
  context-sensitive models for lexical substitution}.
\newblock In \emph{Proceedings of the 13th International Conference on
  Computational Semantics - Long Papers}, pages 271--282, Gothenburg, Sweden.
  Association for Computational Linguistics.

\bibitem[{Hadiwinoto et~al.(2019)Hadiwinoto, Ng, and
  Gan}]{hadiwinoto-etal-2019-improved}
Christian Hadiwinoto, Hwee~Tou Ng, and Wee~Chung Gan. 2019.
\newblock \href {https://doi.org/10.18653/v1/D19-1533} {Improved word sense
  disambiguation using pre-trained contextualized word representations}.
\newblock In \emph{Proceedings of the 2019 Conference on Empirical Methods in
  Natural Language Processing and the 9th International Joint Conference on
  Natural Language Processing (EMNLP-IJCNLP)}, pages 5297--5306, Hong Kong,
  China. Association for Computational Linguistics.

\bibitem[{He et~al.(2020)He, Liu, Gao, and Chen}]{he2020deberta}
Pengcheng He, Xiaodong Liu, Jianfeng Gao, and Weizhu Chen. 2020.
\newblock \href {https://arxiv.org/abs/2006.03654} {Deberta: Decoding-enhanced
  bert with disentangled attention}.
\newblock \emph{arXiv preprint arXiv:2006.03654}.

\bibitem[{Kim et~al.(2021)Kim, Yoo, and Lee}]{kim-etal-2021-self}
Taeuk Kim, Kang~Min Yoo, and Sang-goo Lee. 2021.
\newblock \href {https://doi.org/10.18653/v1/2021.acl-long.197} {Self-guided
  contrastive learning for {BERT} sentence representations}.
\newblock In \emph{Proceedings of the 59th Annual Meeting of the Association
  for Computational Linguistics and the 11th International Joint Conference on
  Natural Language Processing (Volume 1: Long Papers)}, pages 2528--2540,
  Online. Association for Computational Linguistics.

\bibitem[{Levine et~al.(2020)Levine, Lenz, Dagan, Ram, Padnos, Sharir,
  Shalev-Shwartz, Shashua, and Shoham}]{levine-etal-2020-sensebert}
Yoav Levine, Barak Lenz, Or~Dagan, Ori Ram, Dan Padnos, Or~Sharir, Shai
  Shalev-Shwartz, Amnon Shashua, and Yoav Shoham. 2020.
\newblock \href {https://doi.org/10.18653/v1/2020.acl-main.423} {{S}ense{BERT}:
  Driving some sense into {BERT}}.
\newblock In \emph{Proceedings of the 58th Annual Meeting of the Association
  for Computational Linguistics}, pages 4656--4667, Online. Association for
  Computational Linguistics.

\bibitem[{Liu et~al.(2021{\natexlab{a}})Liu, Shareghi, Meng, Basaldella, and
  Collier}]{liu2020self}
Fangyu Liu, Ehsan Shareghi, Zaiqiao Meng, Marco Basaldella, and Nigel Collier.
  2021{\natexlab{a}}.
\newblock \href {https://arxiv.org/pdf/2010.11784.pdf} {Self-alignment
  pretraining for biomedical entity representations}.
\newblock In \emph{Proceedings of the 2021 Conference of the North {A}merican
  Chapter of the Association for Computational Linguistics: Human Language
  Technologies}. Association for Computational Linguistics.

\bibitem[{Liu et~al.(2021{\natexlab{b}})Liu, Vuli{\'c}, Korhonen, and
  Collier}]{liu2021fast}
Fangyu Liu, Ivan Vuli{\'c}, Anna Korhonen, and Nigel Collier.
  2021{\natexlab{b}}.
\newblock \href {https://arxiv.org/abs/2104.08027} {Fast, effective and
  self-supervised: Transforming masked language models into universal lexical
  and sentence encoders}.
\newblock \emph{arXiv preprint arXiv:2104.08027}.

\bibitem[{Liu et~al.(2020)Liu, McCarthy, and Korhonen}]{liu2020towards}
Qianchu Liu, Diana McCarthy, and Anna Korhonen. 2020.
\newblock \href {https://doi.org/10.18653/v1/2020.emnlp-main.333} {Towards
  better context-aware lexical semantics:adjusting contextualized
  representations through static anchors}.
\newblock In \emph{Proceedings of the 2020 Conference on Empirical Methods in
  Natural Language Processing (EMNLP)}, pages 4066--4075, Online. Association
  for Computational Linguistics.

\bibitem[{Liu et~al.(2021{\natexlab{c}})Liu, Ponti, McCarthy, Vuli{\'c}, and
  Korhonen}]{liu2021am2ico}
Qianchu Liu, Edoardo~M Ponti, Diana McCarthy, Ivan Vuli{\'c}, and Anna
  Korhonen. 2021{\natexlab{c}}.
\newblock \href {https://arxiv.org/abs/2104.08639} {Am2ico: Evaluating word
  meaning in context across low-resource languages with adversarial examples}.
\newblock \emph{arXiv preprint arXiv:2104.08639}.

\bibitem[{Liu et~al.(2019)Liu, Ott, Goyal, Du, Joshi, Chen, Levy, Lewis,
  Zettlemoyer, and Stoyanov}]{liu2019roberta}
Yinhan Liu, Myle Ott, Naman Goyal, Jingfei Du, Mandar Joshi, Danqi Chen, Omer
  Levy, Mike Lewis, Luke Zettlemoyer, and Veselin Stoyanov. 2019.
\newblock Roberta: A robustly optimized bert pretraining approach.
\newblock \emph{arXiv preprint arXiv:1907.11692}.

\bibitem[{Loshchilov and Hutter(2019)}]{loshchilov2018decoupled}
Ilya Loshchilov and Frank Hutter. 2019.
\newblock \href {https://openreview.net/forum?id=Bkg6RiCqY7} {Decoupled weight
  decay regularization}.
\newblock In \emph{7th International Conference on Learning Representations,
  {ICLR} 2019, New Orleans, LA, USA, May 6-9, 2019}. OpenReview.net.

\bibitem[{Loureiro and Jorge(2019)}]{loureiro-jorge-2019-language}
Daniel Loureiro and Al{\'\i}pio Jorge. 2019.
\newblock \href {https://doi.org/10.18653/v1/P19-1569} {Language modelling
  makes sense: Propagating representations through {W}ord{N}et for
  full-coverage word sense disambiguation}.
\newblock In \emph{Proceedings of the 57th Annual Meeting of the Association
  for Computational Linguistics}, pages 5682--5691, Florence, Italy.
  Association for Computational Linguistics.

\bibitem[{Mickus et~al.(2020)Mickus, Paperno, Constant, and van
  Deemter}]{mickus-etal-2020-mean}
Timothee Mickus, Denis Paperno, Mathieu Constant, and Kees van Deemter. 2020.
\newblock \href {https://www.aclweb.org/anthology/2020.scil-1.35} {What do you
  mean, {BERT}?}
\newblock In \emph{Proceedings of the Society for Computation in Linguistics
  2020}, pages 279--290, New York, New York. Association for Computational
  Linguistics.

\bibitem[{Mu and Viswanath(2018)}]{mu2017all}
Jiaqi Mu and Pramod Viswanath. 2018.
\newblock \href {https://openreview.net/forum?id=HkuGJ3kCb} {All-but-the-top:
  Simple and effective postprocessing for word representations}.
\newblock In \emph{6th International Conference on Learning Representations,
  {ICLR} 2018, Vancouver, BC, Canada, April 30 - May 3, 2018, Conference Track
  Proceedings}. OpenReview.net.

\bibitem[{Navigli(2009)}]{navigli2009word}
Roberto Navigli. 2009.
\newblock \href {https://dl.acm.org/doi/10.1145/1459352.1459355} {Word sense
  disambiguation: {A} survey}.
\newblock \emph{ACM Computing Surveys}, 41(2):1--69.

\bibitem[{Oord et~al.(2018)Oord, Li, and Vinyals}]{oord2018representation}
Aaron van~den Oord, Yazhe Li, and Oriol Vinyals. 2018.
\newblock \href {https://arxiv.org/pdf/1807.03748.pdf} {Representation learning
  with contrastive predictive coding}.
\newblock \emph{arXiv preprint arXiv:1807.03748}.

\bibitem[{Pedinotti and Lenci(2020)}]{pedinotti-lenci-2020-dont}
Paolo Pedinotti and Alessandro Lenci. 2020.
\newblock \href {https://doi.org/10.18653/v1/2020.coling-main.602} {Don{'}t
  invite {BERT} to drink a bottle: Modeling the interpretation of metonymies
  using {BERT} and distributional representations}.
\newblock In \emph{Proceedings of the 28th International Conference on
  Computational Linguistics}, pages 6831--6837, Barcelona, Spain (Online).
  International Committee on Computational Linguistics.

\bibitem[{Pilehvar and Camacho-Collados(2019)}]{pilehvar2019wic}
Mohammad~Taher Pilehvar and Jose Camacho-Collados. 2019.
\newblock \href {https://doi.org/10.18653/v1/N19-1128} {{W}i{C}: the
  word-in-context dataset for evaluating context-sensitive meaning
  representations}.
\newblock In \emph{Proceedings of the 2019 Conference of the North {A}merican
  Chapter of the Association for Computational Linguistics: Human Language
  Technologies, Volume 1 (Long and Short Papers)}, pages 1267--1273,
  Minneapolis, Minnesota. Association for Computational Linguistics.

\bibitem[{Raganato et~al.(2017)Raganato, Camacho{-}Collados, and
  Navigli}]{Raganato:2017eacl}
Alessandro Raganato, Jos{\'{e}} Camacho{-}Collados, and Roberto Navigli. 2017.
\newblock \href {https://doi.org/10.18653/v1/e17-1010} {Word sense
  disambiguation: {A} unified evaluation framework and empirical comparison}.
\newblock In \emph{Proceedings of EACL 2017}, pages 99--110.

\bibitem[{Raganato et~al.(2020)Raganato, Pasini, Camacho-Collados, and
  Pilehvar}]{raganato-etal-2020-xl}
Alessandro Raganato, Tommaso Pasini, Jose Camacho-Collados, and Mohammad~Taher
  Pilehvar. 2020.
\newblock \href {https://doi.org/10.18653/v1/2020.emnlp-main.584}
  {{XL}-{W}i{C}: A multilingual benchmark for evaluating semantic
  contextualization}.
\newblock In \emph{Proceedings of the 2020 Conference on Empirical Methods in
  Natural Language Processing (EMNLP)}, pages 7193--7206, Online. Association
  for Computational Linguistics.

\bibitem[{Rajaee and Pilehvar(2021)}]{rajaee-pilehvar-2021-cluster}
Sara Rajaee and Mohammad~Taher Pilehvar. 2021.
\newblock \href {https://doi.org/10.18653/v1/2021.acl-short.73} {A
  cluster-based approach for improving isotropy in contextual embedding space}.
\newblock In \emph{Proceedings of the 59th Annual Meeting of the Association
  for Computational Linguistics and the 11th International Joint Conference on
  Natural Language Processing (Volume 2: Short Papers)}, pages 575--584,
  Online. Association for Computational Linguistics.

\bibitem[{Reif et~al.(2019)Reif, Yuan, Wattenberg, Vi{\'{e}}gas, Coenen,
  Pearce, and Kim}]{Reif:2019neurips}
Emily Reif, Ann Yuan, Martin Wattenberg, Fernanda~B. Vi{\'{e}}gas, Andy Coenen,
  Adam Pearce, and Been Kim. 2019.
\newblock \href
  {https://proceedings.neurips.cc/paper/2019/hash/159c1ffe5b61b41b3c4d8f4c2150f6c4-Abstract.html}
  {Visualizing and measuring the geometry of {BERT}}.
\newblock In \emph{Advances in Neural Information Processing Systems 32: Annual
  Conference on Neural Information Processing Systems 2019, NeurIPS 2019,
  December 8-14, 2019, Vancouver, BC, Canada}, pages 8592--8600.

\bibitem[{Reimers and Gurevych(2019{\natexlab{a}})}]{reimers2019sentence}
Nils Reimers and Iryna Gurevych. 2019{\natexlab{a}}.
\newblock \href {https://doi.org/10.18653/v1/D19-1410} {Sentence-{BERT}:
  Sentence embeddings using {S}iamese {BERT}-networks}.
\newblock In \emph{Proceedings of the 2019 Conference on Empirical Methods in
  Natural Language Processing and the 9th International Joint Conference on
  Natural Language Processing (EMNLP-IJCNLP)}, pages 3982--3992, Hong Kong,
  China. Association for Computational Linguistics.

\bibitem[{Reimers and
  Gurevych(2019{\natexlab{b}})}]{reimers-gurevych-2019-sentence}
Nils Reimers and Iryna Gurevych. 2019{\natexlab{b}}.
\newblock \href {https://doi.org/10.18653/v1/D19-1410} {Sentence-{BERT}:
  Sentence embeddings using {S}iamese {BERT}-networks}.
\newblock In \emph{Proceedings of the 2019 Conference on Empirical Methods in
  Natural Language Processing and the 9th International Joint Conference on
  Natural Language Processing (EMNLP-IJCNLP)}, pages 3982--3992, Hong Kong,
  China. Association for Computational Linguistics.

\bibitem[{Ruder(2021)}]{Ruder:2021blog}
Sebastian Ruder. 2021.
\newblock Recent advances in language model fine-tuning.
\newblock \url{http://ruder.io/recent-advances-lm-fine-tuning}.

\bibitem[{Vuli{\'c} et~al.(2020)Vuli{\'c}, Ponti, Litschko, Glava{\v{s}}, and
  Korhonen}]{vulic-etal-2020-probing}
Ivan Vuli{\'c}, Edoardo~Maria Ponti, Robert Litschko, Goran Glava{\v{s}}, and
  Anna Korhonen. 2020.
\newblock \href {https://doi.org/10.18653/v1/2020.emnlp-main.586} {Probing
  pretrained language models for lexical semantics}.
\newblock In \emph{Proceedings of the 2020 Conference on Empirical Methods in
  Natural Language Processing (EMNLP)}, pages 7222--7240, Online. Association
  for Computational Linguistics.

\bibitem[{Wang et~al.(2019)Wang, Pruksachatkun, Nangia, Singh, Michael, Hill,
  Levy, and Bowman}]{Wang:2019superglue}
Alex Wang, Yada Pruksachatkun, Nikita Nangia, Amanpreet Singh, Julian Michael,
  Felix Hill, Omer Levy, and Samuel~R. Bowman. 2019.
\newblock \href
  {https://proceedings.neurips.cc/paper/2019/hash/4496bf24afe7fab6f046bf4923da8de6-Abstract.html}
  {Superglue: {A} stickier benchmark for general-purpose language understanding
  systems}.
\newblock In \emph{Advances in Neural Information Processing Systems 32: Annual
  Conference on Neural Information Processing Systems 2019, NeurIPS 2019,
  December 8-14, 2019, Vancouver, BC, Canada}, pages 3261--3275.

\bibitem[{Wiedemann et~al.(2019)Wiedemann, Remus, Chawla, and
  Biemann}]{wiedemann2019does}
Gregor Wiedemann, Steffen Remus, Avi Chawla, and Chris Biemann. 2019.
\newblock \href {https://arxiv.org/abs/1909.10430} {Does bert make any sense?
  interpretable word sense disambiguation with contextualized embeddings}.
\newblock \emph{arXiv preprint arXiv:1909.10430}.

\bibitem[{Xu et~al.(2020)Xu, Qiu, Zhou, and Huang}]{xu2020improving}
Yige Xu, Xipeng Qiu, Ligao Zhou, and Xuanjing Huang. 2020.
\newblock \href {https://arxiv.org/abs/2002.10345} {Improving bert fine-tuning
  via self-ensemble and self-distillation}.
\newblock \emph{arXiv preprint arXiv:2002.10345}.

\bibitem[{Yan et~al.(2021)Yan, Li, Wang, Zhang, Wu, and Xu}]{yan2021consert}
Yuanmeng Yan, Rumei Li, Sirui Wang, Fuzheng Zhang, Wei Wu, and Weiran Xu. 2021.
\newblock \href {https://arxiv.org/abs/2105.11741} {Consert: A contrastive
  framework for self-supervised sentence representation transfer}.
\newblock \emph{arXiv preprint arXiv:2105.11741}.

\bibitem[{Zhang et~al.(2021)Zhang, He, Liu, Bing, and
  Li}]{zhang-etal-2021-bootstrapped}
Yan Zhang, Ruidan He, Zuozhu Liu, Lidong Bing, and Haizhou Li. 2021.
\newblock \href {https://doi.org/10.18653/v1/2021.acl-long.402} {Bootstrapped
  unsupervised sentence representation learning}.
\newblock In \emph{Proceedings of the 59th Annual Meeting of the Association
  for Computational Linguistics and the 11th International Joint Conference on
  Natural Language Processing (Volume 1: Long Papers)}, pages 5168--5180,
  Online. Association for Computational Linguistics.

\end{thebibliography}
\bibliographystyle{acl_natbib}


\appendix

\section{Appendix}
\label{sec:appendix}

\subsection{AUC Score Tables for Binary Classification Tasks}\label{sec:auc}

\begin{table}[h]
\small
\setlength{\tabcolsep}{2pt}
\centering
\resizebox{0.99\columnwidth}{!}{
\begin{tabular}{lcccccccccc}
\toprule
model$\downarrow$, dataset$\rightarrow$ & WiC &	TSV-1 &	TSV-2 &  TSV-3  \\ 
\midrule
Sentence-BERT & 64.20 & 63.99 &61.81&66.01\\
  \mb & 67.31 & 70.53 & 68.00 & 70.28 \\
 \midrule  
BERT &	71.61 & 62.06 & 59.48& 61.45\\
 \rowcolor{cyan!10}
  + \model & 74.89 & 72.10	&  67.92 & 73.03 \\
  \midrule  
 DeBERTa & 70.58 & 62.11 & 60.51 & 63.25  \\
  \rowcolor{cyan!10}
 + \model & \textbf{76.70} & \textbf{75.44} & \textbf{71.24} & \textbf{75.64} \\
\bottomrule
\end{tabular}
}%
\caption{AUC results for English tasks.}
\label{Table:wic_auc}
\end{table}

\begin{table*}[!t]
\small
\centering
\begin{tabular}{lccccccccccccccccccccccccccccccccc}
\toprule
level$\rightarrow$& \multicolumn{4}{c}{XL-WiC} &&  \multicolumn{4}{c}{AM2iCo} \\

 \cmidrule{2-5}\cmidrule{7-10} 
  model$\downarrow$, language$\rightarrow$ & \zh* & \ko* & \hr & \et & & \zh &	\ka & \ja &	\ar \\ 
   \midrule 
BERT & 80.97 & 75.17 & 67.79 & 62.72 && 68.06 &	64.50 & 69.32 &	68.98 \\
 \rowcolor{cyan!10}
  + \model & \textbf{83.39} & \textbf{80.44} & \textbf{76.80} &\textbf{64.62} && \textbf{69.23} & \textbf{65.57} & \textbf{72.86} & \textbf{69.27} \\
\bottomrule
\end{tabular}
\caption{AUC results for multilingual and cross-lingual tasks. }
\label{Table:xl_auc}
\end{table*}

Following prior work, we reported accuracy in the main text. However, the threshold for TRUE/FALSE classification needs to be tuned on dev set. We thus report the AUC scores in \Cref{Table:wic_auc,Table:xl_auc} which does not require  tuning of any hyperparameter. The AUC scores demonstrate the same trend as accuracy scores.

\subsection{Pretrained Encoders Details\label{sec:encoder}}

For a full listing of HuggingFace model links and number of parameters for each model, see \Cref{Table:pretrained_encoder_details}.

\begin{table*}[!ht] 
\small
\centering
\begin{tabular}{lll}
\toprule
model & \#param & URL \\
\midrule
BERT & 110M & \url{https://huggingface.co/bert-base-uncased} \\
RoBERTa & 110M & \url{https://huggingface.co/roberta-base} \\
DeBERTa & 138M& \url{https://huggingface.co/microsoft/deberta-base} \\
mBERT & 168M & \url{https://huggingface.co/bert-base-multilingual-uncased} \\
BERT (\zh) & 103M& \url{https://huggingface.co/bert-base-chinese} \\
BERT (\ko) & 118M &  \url{https://huggingface.co/kykim/bert-kor-base} \\
\bottomrule
\end{tabular}
\caption{A listing of HuggingFace URLs of all pretrained models used in this work.}
\label{Table:pretrained_encoder_details}
\end{table*}

\subsection{Hyperparameter Optimisation}\label{sec:hyperparameters}
\Cref{Table:search_space} shows a full listing of the hyperparameters (and their search space). As said in main text, hyperparameters remain as the same as  set in prior work of \citet{liu2021fast}, except for random span masking rate and dropout rate. 

\begin{table*}[!ht] 
\small
\centering
\begin{tabular}{lr}
\toprule
hyperparameters & search space \\
\midrule
learning rate & \{\texttt{1e-5}, \texttt{2e-5}$^\ast$,\texttt{3e-5}\}\\
 batch size & 200 \\
 training epochs & \{1$^\ast$, 2, 3, 4\} \\
  training data size & \{5k, 10k$^\ast$, 20k, 30k, 40k, 50k\} \\
\texttt{max\_seq\_length} of tokeniser & 50 \\
$\tau$ in \Cref{eq:infonce} & \{0.02, 0.03, 0.04$^{\ast}$, 0.05, 0.06\} \\
random span masking rate (BERT)  & \{0, 1, 5, 10$^{\ast}$, 15\} \\
random span masking rate (RoBERTa)  & \{0$^{\ast}$, 1, 5, 10 15\} \\
random span masking rate (DeBERTa)  & \{0, 1$^\ast$, 5, 10, 15\} \\
dropout rate (BERT) & \{0.1, 0.2, 0.3, 0.4$^\ast$, 0.5, 0.6\}\\
dropout rate (RoBERTa)  & \{0.1, 0.2, 0.3$^\ast$, 0.4, 0.5, 0.6\} \\
dropout rate (DeBERTa)  & \{0.1, 0.2, 0.3$^\ast$, 0.4, 0.5, 0.6\} \\
\bottomrule
\end{tabular}
\caption{Hyperparameters along with their search grid. $\ast$ marks the values used to obtain the reported results. The hparams without any defined search grid are adopted directly from \citet{liu2020self}.}
\label{Table:search_space}
\end{table*}

\subsection{Sensitivity to Training Corpora}
To test the robustness of the model to different corpora, we individually sampled five sets of 10k raw sentences and found only minor difference when fine-tuning on them ($\approx0.003$ standard deviation for BERT +\model and $\approx0.001$ for DeBERTa +\model). We also tested with fine-tuning with strictly `in-domain' data, i.e., raw sentences (w/o labels) sampled from the training sets of WiC tasks, but found no substantial difference when comparing to fine-tuning on Wikipedia texts.

\subsection{Software and Hardware Dependencies}
Our experiments are implemented with PyTorch and Huggingface Transformers. For PyTorch training, Automatic Mixed Precision (AMP)\footnote{\url{https://pytorch.org/docs/stable/amp.html}} is turned on. The hardware configuration is listed in \Cref{Table:hardware}. \model training on this machine takes $\approx30$ seconds.

\begin{table}[h] 
\small
\setlength{\tabcolsep}{1pt}
\centering
\begin{tabular}{lr}
\toprule
hardware & specification \\
\midrule
RAM & 128 GB \\
CPU & AMD Ryzen 9 3900x 12-core processor × 24  \\
 GPU & NVIDIA GeForce RTX 2080 Ti (11 GB) $\times$ 2\\
\bottomrule
\end{tabular}
\caption{Hardware specifications of the used machine.}
\label{Table:hardware}
\end{table}
\end{document}